\documentclass{article}

\PassOptionsToPackage{numbers, sort&compress}{natbib}

\usepackage[final]{neurips_2021}

\usepackage[utf8]{inputenc} 
\usepackage[T1]{fontenc}    
\usepackage{booktabs} 
\usepackage{url,amsthm,amsmath,amssymb,amsfonts}
\usepackage{xcolor}
\usepackage{caption}
\usepackage{subcaption}
\usepackage{algorithm,algpseudocode}
\usepackage{tikz}
\usepackage{bbm}
\usepackage{rbase} 
\usepackage{paralist}
\usepackage{enumitem}
\definecolor{hcitecolor}{RGB}{40,40,160}
\usepackage[colorlinks=true,citecolor=hcitecolor, linkcolor=black]{hyperref}
\usepackage{wrapfig}
\usepackage{graphicx}
\usepackage{pgfplots}
\usepackage[all,2cell,ps,cmtip]{xy}
\usepackage{tikz}
\usepackage{caption}
\usepackage{microtype}
\usepackage{setspace}
\usepackage{soul}

\usepackage{mdframed}
\usepackage{ifthen}
\usetikzlibrary{shadows}
\usetikzlibrary{matrix, calc, positioning, arrows,shapes,backgrounds, arrows.meta, quotes,fit}
\usepackage{gensymb,mathtools,textcomp,xspace,grffile,rbase,soul}
\usepackage{siunitx}
\usepackage{etoolbox}
\robustify\bfseries

\newtheorem{theorem}{Theorem}

\newtheorem{definition}{Definition}

\renewcommand{\sm}[1]{\m{\smash{#1}}}
\newcommand{\neuron}{\mathfrak{n}}
\newcommand{\rfd}{\mathcal{G}}
\renewcommand{\Aut}{\mathop{\mathrm{Aut}}}
\newcommand{\ftm}{B}
\newcommand{\Sm}{\Sbb_m}
\newcommand{\Sk}{\Sbb_k}

\newcommand {\ssseq}[2]{\ensuremath{#1_1\ldots #1_{#2}}}
\newcommand{\dn}{{\downarrow}}
\newcommand{\up}{{\uparrow}}

\algnewcommand\algorithmicinput{\textbf{Input:}}
\algnewcommand\INPUT{\item[\algorithmicinput]}
\algnewcommand\algorithmicretn{\textbf{Return:}}
\algnewcommand\RETURN{\item[\algorithmicretn]}

\newcommand{\changed}[1]{#1}

\newlength{\sectionpadding}
\newlength{\figpadding}
\newlength{\subsectionpadding}
\title{Autobahn: Automorphism-based Graph Neural Nets}

\author{
	Erik H. Thiede,\textsuperscript{1*} Wenda Zhou,\textsuperscript{12\dag} Risi Kondor\textsuperscript{134\ddag}
\\
\textsuperscript{1} Center for Computational Mathematics, Flatiron Institute, New York NY 10010
\\
\textsuperscript{2} Center for Data Science, New York University, New York NY 10011
\\
\textsuperscript{3} Department of Computer Science, University of Chicago, Chicago IL 60637
\\
\textsuperscript{4} Department of Statistics, University of Chicago, Chicago IL 60637
\\
\textsuperscript{*}ehthiede@flatironinstitute.org, \textsuperscript{\dag}wz2247@nyu.edu, \textsuperscript{\ddag}risi@uchicago.edu
}
\begin{document}

\maketitle

\begin{abstract}
    We introduce Automorphism-based graph neural networks (Autobahn), a new family of graph neural networks. In an Autobahn, we decompose the graph into a collection of subgraphs and apply local convolutions that are equivariant to each subgraph's automorphism group. Specific choices of local neighborhoods and subgraphs recover existing architectures such as message passing neural networks. Our formalism also encompasses novel architectures: as an example, we introduce a graph neural network that decomposes the graph into paths and cycles. The resulting convolutions reflect the natural way that parts of the graph can transform, preserving the intuitive meaning of convolution without sacrificing global permutation equivariance. We validate our approach by applying Autobahn to molecular graphs, where it achieves results competitive with state-of-the-art message passing algorithms.
\end{abstract}

\vspace{\sectionpadding}
\section{Introduction}
\vspace{\sectionpadding}
The successes of artificial neural networks in domains such as computer vision and natural language processing
have inspired substantial interest in developing neural architectures on graphs.
Since graphs naturally capture relational information, graph-structured data appears in a myriad of fields.
However, working with graph data raises new problems.
Chief among them is the problem of graph isomorphism:
for the output of our neural network to be reliable, 
it is critical that the network gives the same result independent of trivial changes in graph representation 
such as permutation of nodes.

Considerable effort has gone into constructing neural network architectures that obey this constraint\cite{scarselli2008graph,perozzi2014deepwalk,HenaffLeCun2015,Defferrard2016, bronstein2017geometric}.
Arguably, the most popular approach is to construct \emph{Message-Passing Neural Networks (MPNNs)}\cite{KipfWelling2017,Gilmer2017}.
In each layer of an MPNN, every node aggregates the activations of its neighbors in a permutation invariant manner
and applies a linear mixing and nonlinearity to the resulting vector.
While subsequent architectures have built on this paradigm, e.g., by improving activations on nodes and graph edges \cite{xu2018powerful, hu2019strategies},
the core paradigm of repeatedly pooling information from neighboring nodes has remained.
These architectures are memory efficient, intuitively appealing, and respect the graph's symmetry under permutation of its nodes.
However, practical results have shown that they can oversmooth signals \cite{li2019deepgcns} and theoretical work 
has shown that they have trouble distinguishing certain graphs and counting substructures \cite{xu2018powerful,arvind2020weisfeiler,chen2020can,garg2020generalization}.
Moreover, MPNNs do not use the graph's topology to its fullest extent.
For instance, applying an MPNN to highly structured graphs such as grid graphs does not recover powerful known architectures such as convolutional neural networks.
It is also not clear how to best adapt MPNNs to families of graphs with radically different topologies:
MPNNs for citation graphs and molecular graphs are constructed in largely the same way.
This suggests that it should be possible to construct more expressive graph neural networks by directly leveraging the graph's structure.

In this work, we introduce a new framework for constructing graph neural networks, \textit{Automorphism-based Neural Networks} (Autobahn).
Our research is motivated by our goal of designing neural networks that can learn the properties of small organic molecules accurately enough to make a significant contribution to drug discovery and materials design \cite{korotcov2017comparison,bock2019review,haghighatlari2019advances,pollice2021data}. 
The properties of these molecules depend crucially on multi-atom substructures, making the difficulties MPNNs have in recognizing substructures a critical problem.
We realized that we could circumvent this limitation by making graph substructures \emph{themselves} the fundamental units of computation.
The symmetries of our substructures then inform the computation.
For instance, the benzene molecule forms a ring of six atoms and is a common subunit in larger molecules.
On this ring we have a very natural and mathematically rigorous notion of convolution: convolution on a  one-dimensional, periodic domain.
This symmetry is encoded by a graph's automorphism group:
the group that reflects our substructures' internal symmetries.
Our networks directly leverage the automorphism group of subgraphs to construct flexible, efficient neurons.
Message passing neural networks arise naturally in this framework 
when the substructures used are local star graphs,
and applying Autobahn to grid graphs can recover standard convolutional architectures
such as steerable CNNs.
More generally, Autobahn gives practitioners the tools 
to build bespoke graph neural networks whose substructures reflect their domain knowledge.
As an example, in Section~\ref{sec:experiments}, we present a novel architecture 
outside of the message-passing paradigm that nevertheless achieves performance competitive with state-of-the-art MPNNs on molecular learning tasks.

\vspace{\sectionpadding}
\section{Graph Neural Networks}\label{sec:gnn_perm_equiv}
\vspace{\sectionpadding}
Neural Networks operate by composing several learned mappings known as ``layers''.
Denoting the $\ell$'th layer in the network as $\phi_\ell$, the functional form of a neural network $\Phi$ can be written as
\begin{equation*}
    \Phi=\phi_L \circ \phi_{L-1}\circ \ldots \circ \phi_1.  
\end{equation*}
Each layer is typically constructed from a collection of parts, the titular ``neurons''.
We denote the $i$'th neuron in the $\ell$'th layer as \sm{\neuron^\ell_i},
and denote its output (the ``activation'' of the neuron) as \sm{f^\ell_i}.
Architectures differ primarily in how the neurons are constructed and how their inputs and outputs are combined.

When constructing a neural network that operates on graph data, care must be taken to preserve the input graphs' natural symmetries under permutation.
Let \m{\Gcal} be a graph with node set \m{\cbrN{\sseq{v}{n}}}, 
adjacency matrix \sm{A\tin\RR^{n\times n}} and 
\m{d}-dimensional node labels  \sm{b_i\in\RR^d} 
stacked into a matrix \sm{B\tin\RR^{n\times d}}. 
Permuting the numbering of the nodes of \m{\Gcal} by some 
permutation \m{\sigma\colon \cbr{\oneton{n}}\to\cbr{\oneton{n}}} transforms
\begin{equation}\label{eq: Aact}
A\mapsto A^\sigma\hspace{80pt} A^\sigma_{i,j}=A_{\sigma^{-1}(i),\sigma^{-1}(j)},
\end{equation}
and 
\begin{equation}\label{eq: Bact}
\ftm{}\mapsto \ftm{}^\sigma\hspace{100pt} \ftm^\sigma_{i,n}=\ftm_{\sigma^{-1}(i),n}.
\end{equation}
This transformation does not change the actual topology of \m{\Gcal}.
Consequently, a fundamental requirement on graph neural networks is that they be invariant
with respect to such permutations.

\vspace{\subsectionpadding}
\subsection{Message-Passing Neural Networks}
\vspace{\subsectionpadding}
\begin{figure}
  \begin{center}
    \includegraphics[width=0.6\textwidth]{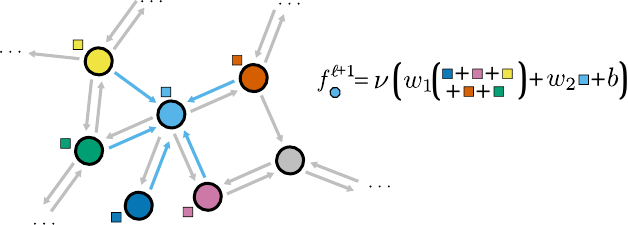}
  \end{center}
  \caption{
           Visualization of a single neuron of a simple message passing neural network.
           Each node aggregates the features from neighboring nodes using a permutation-invariant operation
           (we use summation for simplicity), applies learned weight matrices and biases and finally a nonlinearity.
           }
                      \vspace{1.0\figpadding}
  \label{fig:mpnn}
\end{figure}
Message-passing neural networks (MPNNs) have emerged as the dominant paradigm for constructing neural networks on graphs.
Every neuron in an MPNN corresponds to a single node in the graph.
Neurons aggregate features from neighboring nodes by passing them through 
a permutation invariant function.
They then combine the result with the node's original message, and pass the result through a learned linear function and nonlinearity (Figure~\ref{fig:mpnn})\cite{xu2018powerful}.
Since information is transmitted over the graph's topology,
MPNNs are not confounded by permutations of the graph.
Specific architectures may differ in the details of the aggregation function \cite{KipfWelling2017,Duvenaud2015,hamilton2017inductive,xu2018powerful},
may include additional terms to account for edge features \cite{xu2018powerful,hu2019strategies},
may use complex transformations to construct the node features encoding local structure \cite{alsentzer2020subgraph,bouritsas2020improving},
or may augment the graph with additional nodes \cite{JunctionTreeVAE_ICML2018,fey2020hierarchical}.

\vspace{\sectionpadding}
\section{Permutation Equivariance}\label{sec:perm_equivariance}
\vspace{\sectionpadding}
As discussed in the Introduction, MPNNs have fundamental limits to their expressiveness.
To construct more powerful neural networks, we turn to the general formalism of group equivariant networks \cite{KondorTrivedi2018,Cohen2016,Cohen2017,CohenCNNhomo}. 
Our desire that permutations of the input graph leave our network's output unaffected is formalized by the notion of group-invariance.
Let us assume that our input data lives in a space $X$ that is acted on by a group $G$, and for all $g \in G$ denote the associated group action on $X$ by $T_g$. 
The invariance constraint amounts to requiring:
\begin{equation}
    \Phi = \Phi \circ T_g \hspace{100pt} \forall g \in G.
    \label{eq:defn_invariance}
\end{equation}
One way to satisfy this would be to require that each layer \m{\phi_\ell} 
be fully invariant to $G$. 
However, in practice this condition can be extremely restrictive.
For this reason, networks commonly use group \emph{equivariant} layers.
Let  $X$ and $Y$ be the input and output spaces of $\phi_\ell$, 
with group actions $T_g$ and $T'_g$, respectively.
We say $\phi_\ell$ is equivariant to $G$ if it obeys
\begin{equation}
    T'_g \circ \phi_\ell  = \phi_\ell \circ T_g \hspace{90pt} \forall g \in G.
    \label{eq:defn_equiviariance}
\end{equation}
This condition is weaker than invariance: we recover invariance when $T_g'$ maps every element of $Y$ to itself.
Moreover, it is simple to show that the composition of two equivariant layers is also equivariant. 
Consequently, in all layers but the last we can enforce the weaker condition of equivariance
and merely enforce invariance in the final layer. In the case of graph neural networks, the relevant group is the group of permutations: \m{\Sn} (called the symmetric group of degree \m{n}).

\vspace{\subsectionpadding}
\subsection{Permutation-equivariant networks}\label{ssec:sn_equivariant_layers}
\vspace{\subsectionpadding}
Recent work has developed a generic recipe for constructing group equivariant networks.
In this formalism, any object that transforms under a group action is treated as a function on the group \cite{KondorTrivedi2018} (see Section~1 in the supplement for a brief review).
This allows all layers equivariant to the group to be treated using the same formalism, independently of how inputs and outputs transform.
In particular, it can be shown that the only group-equivariant linear operation possible is a generalized notion of group convolution.
For discrete groups, this convolution can be written as:
\begin{equation}
    \left(f \ast w\right)(u)  = \sum_{v \in G} f(u v^{-1} )\, w (v ).
    \label{eq:generic_convolution}
\end{equation}
To construct an equivariant neuron, we apply \eqref{eq:generic_convolution} to convolve our input activation $f^{\ell-1}$ with a learned weight function $w$, add a bias, and then apply a fixed equivariant nonlinearity, $\nu$.
Applying this approach to specific groups recovers the standard convolutional layers used in convolutional neural networks (CNNs).
For instance, applying~\eqref{eq:generic_convolution} to the cyclic group of order $n$ gives
\begin{equation}
    (f \ast w)_j = \sum_{k=0}^n f(r^{j-k})\, w(r^{k}),
    \label{eq:one_dimensional_conv}
\end{equation}
where $r$ is the group element corresponding to rotation by $360 / n$ degrees.
Similarly, applying~\eqref{eq:generic_convolution} to one-dimensional or two-dimensional discrete translation groups 
recovers the standard convolutions used for image processing.

Instantiating this theory with the symmetric group has been successfully used to construct permutation equivariant networks for learning tasks defined on sets \cite{DeepSets} 
and for graph neural networks 
\cite{thiede2020general,maron2018invariant,maron2019universality,maron2019provably}. However, enforcing equivariance to all permutations can be very restrictive.
As an example, consider a layer whose domain and range transform according to~\eqref{eq: Bact} with $c_{in}$ incoming and $c_{out}$ outgoing channels.
Let $w_1, w_2 \tin \RR^{c_{out} \times c_{in}}$ be learned weight matrices, and $f_j \tin \RR^{c_{in}}$ be the vector of incoming activations corresponding to node $j$.
The most general possible convolution is then \cite{DeepSets}
\begin{equation}
    \left(f \ast w\right)_i = w_1 f_i + w_2 \sum_{j=1}^n f_j,
    \label{eq:first_order_conv}
\end{equation}
but this is quite a weak model as activations from different nodes only interact through their sum. 
Consequently, relationships between nodes can only be captured in aggregate.

To address this fundamental limitation, several recent works considered improving the expressivity of MPNNs by defining higher order activations corresponding to pairs, triplets, or, in general, $k$--tuples of nodes \cite{kondor2018covariant,maron2019provably,thiede2020general}. Mathematically, this requires considering not just \rf{eq: Aact} and \rf{eq: Bact},  but the action of the 
symmetric group on \m{k}'th order tensors, 
$A^\sigma_{i_1,\ldots,i_k}\!=A_{\sigma^{-1}(i_1),\ldots,\sigma^{-1}(i_k)}$.
However, this can be prohibitively expensive for many nodes. For instance, organic chemistry depends crucially on the existence of aromatic rings, typically of six atoms. Manipulating sixth order tensors would be extremely costly.

\vspace{\sectionpadding}
\section{Permutation-Equivariant Neurons using Automorphism}\label{sec:eq_and_automorphism}
\begin{figure}[h]
    \begin{subfigure}[]{.23\columnwidth}
    \centering
    \includegraphics[width=\columnwidth]{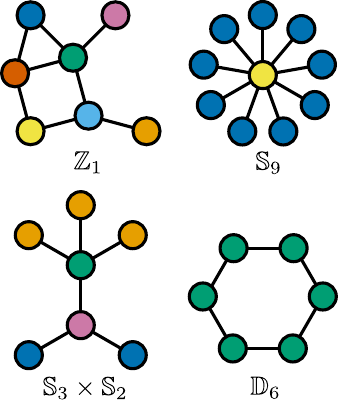}
    \caption{
    }
    \label{fig:automorphism_examples}
    \end{subfigure}
\hfill
    \begin{subfigure}[]{.67\columnwidth}
    \centering
    \centerline{\includegraphics[width=\columnwidth]{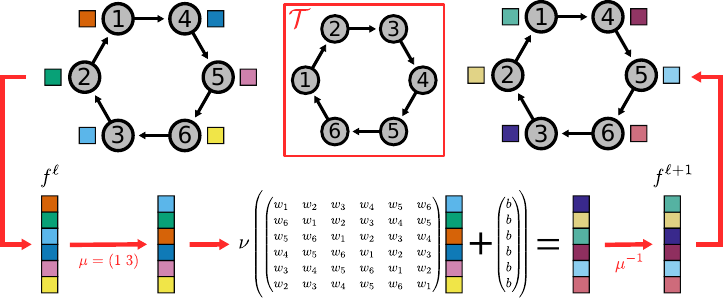}}
    \caption{
    }
    \label{fig:automorphism_layer}
    \end{subfigure}
    \caption{
        Figures visualizing the automorphism group of a graph and its use in graph learning. 
        (a) Four graphs and their automorphism groups.
        In each graph, nodes in the same orbit of the graph's automorphism group are the same color.
        (b) A neuron constructed by applying Algorithm~\ref{algo:autoconv} to a cyclic directed graph.
        We consider the simplified setting where the layer operates only on a single channel of node features.
        Note that the matching step can only be accomplished up to an element in the graph's automorphism group: the cyclic group of order six, $\mathbb{C}_6$.
        \vspace{\figpadding}
    }
\end{figure}

\vspace{\sectionpadding}
The key theoretical idea underlying our work is that to construct more flexible neural networks, we can  exploit the graph topology itself.
In particular,
the local adjacency matrix itself can be used to judiciously break permutation symmetry, allowing us to identify nodes up to the symmetries of $A$.
Letting $\Gcal$ be a graph as before, the automorphism group $\Aut(\Gcal)$ is: 
\begin{equation}
    \Aut(\Gcal) = \left\{\,\sigma \tin \Sn\, |\, A^\sigma\nts = A\,  \right\},
\end{equation}
where \sm{A^\sigma} is defined as in \rf{eq: Aact}. 
Figure~\ref{fig:automorphism_examples} shows the automorphism group of four example graphs.

If $\Gcal$ and $\Gcal'$ are two isomorphic graphs, then each node or edge in $\Gcal'$ can always be matched to a node or edge in $\Gcal$ \emph{up to a permutation in} $\Aut(\Gcal)$.
If every graph our network observed was in the same isomorphism class, we could construct a neuron in a permutation-equivariant neural network by matching $\Gcal$ to a template graph $\Tcal$ and convolving over $\Aut(\Tcal)$.
We give pseudocode for such a neuron, which we call an ``Automorphism-based neuron,'' in Algorithm~\ref{algo:autoconv}.
Note that although the convolution itself is only equivariant to $\Aut(\Tcal)$, the entire neuron is permutation-equivariant.
A formal proof of equivariance is given in Section~2 of the Supplement.

\begin{algorithm}[t]
    \caption{Automorphism-based Neuron
    }
    \begin{algorithmic}[1]
        \INPUT
        \Statex $f^{\ell-1}$  \Comment Incoming activation associated with $\Gcal$
        \Statex $A_\Gcal$  \Comment Adjacency matrix of $\Gcal$
        \Statex $A_\Tcal$  \Comment Adjacency matrix of the saved template graph.
        \State Find $\mu\in \Sn$ such that $A_\Gcal^{\mu} = A_\Tcal$.
        \State $(f^{\ell-1})^{\mu} \gets T_\mu (f^{\ell-1})$ \Comment Apply $\mu$ to incoming activation.
        \State $(f^{\ell}){}^{\mu} \gets \nu \left( (f^{\ell-1} )^{\mu} \ast w + b \right) $ \Comment Convolution is over $\Aut(\Tcal)$.
        \State $f^{\ell} \gets T_{\mu^{-1}}' ((f^{\ell})^{\mu})$ \Comment Map output to original ordering
        \RETURN $f^{\ell}$
    \end{algorithmic}
\label{algo:autoconv}
\end{algorithm}
For a given input and output space, neurons constructed using Algorithm~\ref{algo:autoconv} are more flexible than a neuron constructed using only $\Sn$ convolution.
As an example, consider a neuron operating on graphs isomorphic to the top left graph in Figure~\ref{fig:automorphism_examples} and whose input and output features are a single channel of node features.
An $\Sn$-convolution would operate according to~\eqref{eq:first_order_conv}:
Each output node feature would see only the corresponding input feature and an aggregate of all other node features.
In contrast, Algorithm~\ref{algo:autoconv} completely canonicalizes the graph since this particular graph has no non-trivial automorphisms.
Consequently, we do not need to worry about enforcing symmetry and we can simply run a fully connected layer: a richer representation, since it does not require the same degree of "parameter sharing" as \rf{eq:first_order_conv}.
Similarly, Figure~\ref{fig:automorphism_layer} depicts an analogous neuron constructed in the presence of cyclic graph symmetry.  
In this case, the matching can be performed up to a cyclic permutation, so convolution must be equivariant to the graph's automorphism group, $C_6$.

This general strategy was first described by de Haan et al.~\cite{dehaan2020natural}.
While the increased flexibility was noted, the authors also observed rightfully that this strategy is not directly practical.
Most graph learning problems consist either of many graphs from different isomorphism classes or a single large graph that is only partially known.
In the former case, we would have to construct one network for every isomorphism class and each network would only see a fraction of the data.
In the latter, the isomorphism class of the network would be unknown.

\vspace{\sectionpadding}
\section{Autobahn}\label{sec:autobahn}
\vspace{\sectionpadding}
To motivate our approach we consider MPNNs from the perspective of Algorithm~\ref{algo:autoconv}.
Looking at Figure~\ref{fig:mpnn} we see that the message-passing procedure itself forms a star graph, where the leaves of the star correspond to the neighboring nodes.
The automorphism group of a star graph is the set of all
permutations that swap the star's leaves.
This is precisely the group structure of a single MPNN neuron.
Since we apply a permutation-invariant aggregation function, 
if we were to permute a neuron's input data between the leaves of the star, the MPNN neuron would be unaffected.
However, permuting the input features for the central node with one of its neighbors would ``break'' the MPNN.

This suggests a natural generalization of MPNNs.
In every layer, we decompose a graph into a collection of subgraphs known as \emph{local graphs} that are isomorphic to a pre-selected template graph. 
Although in worst case this may be polynomially expensive, for ``real world'' graphs we expect that it can be solved efficiently.  In particular, one can leverage the  well-developed literature of efficient search heuristics  \cite{cordella2001improved,cordella2004sub,junttila2007engineering,han2013turboiso}.
We next construct a permutation-equivariant neuron on each local graph, denoting the local graph of neuron $\neuron_j^\ell$ as $\rfd_j^\ell$.
Each neuron operates by aggregating information from overlapping local graphs and 
then applying Algorithm~\ref{algo:autoconv} to the result.
Because of the important role played by convolutions over subgraphs' automorphism groups, we refer to the resulting networks as 
Automorphism-based Neural Networks, or \emph{Autobahns}.

MPNNs are not the only commonly used algorithm that can be recovered by the Autobahn formalism.
In Section~3 in the supplement we show that using a grid graph template on a larger grid graph recovers steerable CNNs \cite{Cohen2017}\footnote{
Standard CNNs further break the symmetry of the grid graph by introducing a notion of up/down and left/right.
However, if we introduce a notion of edge ``color'' to distinguish horizontal from vertical edges and extend our definition of automorphism
to include color, Autobahn can recover CNNs as well: see Section~3 in the Supplement.}.
The fact that Autobahn networks naturally recover these architectures when MPNNs do not suggests that considering the local automorphism group is a productive direction for incorporating graph structure.

Our work builds on a considerable body of literature on constructing neural networks with higher-order activations \cite{kondor2018covariant,morris2019weisfeiler,vignac2020building,dehaan2020natural}.
Arguably, our work is most closely related to natural graph networks (NGNs), the formalism proposed in~\cite{dehaan2020natural}.
However, NGNs associate neurons with neighborhoods of individual nodes and edges, making the direct use of automorphic convolutions impractical for generic graphs.  
The specific architecture proposed in~\cite{dehaan2020natural} instead combined multiple message passing networks, each applied to a local neighborhood of the graph.
In contrast, Autobahn associates neurons with subgraphs instead of specific nodes or edges.
To our knowledge, this is the first work to explicitly consider constructing neurons equivariant to the automorphism group of subgraphs.
This additional flexibility allows us to build practical networks using the convolutions described in Section~\ref{sec:eq_and_automorphism}

In the discussion that follows, we give a generic treatment of each step in the network, followed by a full description of an Autobahn layer.
We do not specify a specific form for the activations: they may correspond to individual nodes, edges, hyper-edges, or be delocalized over the entire local graph.
This is in keeping with our philosophy of giving a flexible recipe that practitioners can tailor to their specific problems using their domain knowledge.

\vspace{\subsectionpadding}
\subsection{Convolutions using the Automorphism Group}
\vspace{\subsectionpadding}

The convolutions in Autobahn proceed by applying Algorithm~\ref{algo:autoconv} to each neuron's \emph{local} graph.
The precise form of the convolution will depend on how the activation transforms under permutation.
In~\cite{KondorTrivedi2018} it was observed that for any compact group, one could construct the appropriate notion of group convolution
by expressing the activation in the group's Fourier space and applying the noncommmutative generalization of the convolution theorem. 
Subsequent work has lead to software libraries that convolve over arbitrary finite groups \cite{finzi2021practical}.
Moreover, for specific groups simple convolutions are either known or intuitive to derive.
For instance, in Section~\ref{sec:experiments}, our architecture uses directed cycle and path graphs as templates.  
There, group convolutions can be performed using well-known one-dimensional convolutions such as~\eqref{eq:one_dimensional_conv}.

\vspace{\subsectionpadding}
\subsection{Transferring information across neurons}
\vspace{\subsectionpadding}
To build a rich representation of the structure graph we must be able to transmit information between neurons operating on different local graphs.  
In MPNNs and CNNs, each neuron pools information into the central node before transmitting to its neighbors.  This simplifies the task of transmitting information to other neurons, as the output of neuron $\neuron_j^\ell$ becomes a node feature for neuron $\neuron_k^{\ell+1}$.
However, this strategy does not necessarily work for the neurons in Autobahn: our local graphs may not have a central node.
Even if they do, collapsing each neuron's output into a single node could limit our network's expressivity, as it prevents neurons from transmitting equivariant information such as the hidden representations of multiple nodes or (hyper)edges. 

Instead, we observe that any part of an activation that corresponds to nodes shared between two local graphs can be freely copied between the associated neurons.
To transmit information from \sm{\neuron_j^{\ell-1}} to \sm{\neuron_j^{\ell}}, we define two operations, narrowing and promotion, that extend this copying procedure to arbitrary activations.
Narrowing compresses the output of \sm{\neuron_j^{\ell-1}} into the intersection of the two local graphs, and promotion expands the result to the local graph of \sm{\neuron_j^\ell}.

\begin{figure}[t]
    \centering
    \includegraphics[width=1.0\columnwidth]{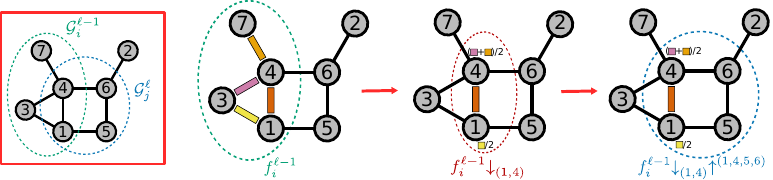}
    \caption{
        Example demonstrating how narrowing and promotion transfer information between local graphs.
        For concreteness, we assume the incoming activation \sm{f_i^{\ell-1}} is a collection of edge features.
        We first narrow to the nodes shared between \sm{\rfd_i^{\ell-1}} and \sm{\rfd_j^\ell}.
        For the edge features used here, this corresponds to averaging over nodes 3 and 7;
        the edge inside the restriction is simply copied. 
        The results are then placed into the appropriate position in the local graph of the output.
        \vspace{\figpadding}
    }
    \label{fig:narrowing_and_promotion}
\end{figure}

To ensure that narrowing and promotion are correctly defined for all activations regardless of their specific group actions, 
we employ the formalism from~\cite{KondorTrivedi2018}, 
where activations are treated as functions on the symmetric group.
Specifically, the input activation and narrowed activation on $m$ and $k$ nodes respectively are identified with functions from $\Sm \to \RR$ and $\Sk \to \RR$. 
We discuss special cases after the definitions, and depict a specific example operating on edge features in Figure~\ref{fig:narrowing_and_promotion}.

\vspace{\subsectionpadding}
\subsubsection{Narrowing}
\vspace{\subsectionpadding}
Narrowing takes an activation \m{f} that transforms under permutation of a 
given set of \m{m} nodes and converts it into a function that transforms only with respect to
a subset of $k\<<m$ nodes, $\cbrN{v_{i_1}, \ldots, v_{i_k}}$.
To construct our narrowed function, we apply an arbitrary permutation 
that ``picks out'' the nodes indexed by $\cbrN{\sseq{i}{k}}$ by sending them to the first $k$ positions.  (Note this implicitly orders these nodes.)
Subsequent permutations of  $\cbrN{\oneton{k}}$ permute our specially chosen nodes amongst each other and permutations of $\cbrN{k+1,\ldots,m}$ permute the other, less desirable nodes.
Narrowing exploits this to construct a function on $\Sk$: we apply the corresponding group element in $\Sk$ to the first $k$ positions, average over all permutations of the last $m-k$, and read off the result.

\begin{definition}\label{defn:narrowing}
    Let $\left(i_1, \ldots, i_k\right)$ be an ordered subset of \m{\cbrN{\oneton{m}}}
    and $t$ be an (arbitrarily chosen) permutation such that 
    \begin{equation}
        t(i_p) = p \hspace{80pt} \forall p\in{1,\ldots,k}.
    \end{equation}
    For all $u\in \Sk$ let $\acute{u}\in \Sm$ be the permutation that applies $u$ to the first $k$ elements and for all $s \in \Sbb_{m-k}$ let $\grave{s} \in \Sm$ be the permutation that applies $s$ to the last $m-k$ elements. 
    Given \m{f\colon\Sm\to\RR^d} we define the \df{narrowing} of \m{f} to \m{(\sseq{i}{k})} as the function:
    \begin{equation}
    f\dn_{(\ssseq{i}{k})}(u)= {(m-k)!}^{-1}\hspace{-5pt} 
    \sum_{s\in \Sbb_{m-k}} f(\acute{u} \grave{s} t).
    \end{equation}
    \vspace{-10pt}
\end{definition}

Narrowing obeys a notion of equivariance for permutations restricted to the local graph.
Let $\sigma$ be a permutation that sends \m{\cbr{\sseq{i}{k}}} to itself.
Narrowing then obeys:
\begin{equation*}
(f\dn_{(\ssseq{i}{k})})^{\sigma'}=(f^{\sigma})\dn_{(\ssseq{i}{k})},
\end{equation*} 
where $\sigma'$ is the permutation in $\Sk$ obeying:
\begin{equation}
    \sigma' (p) = q \qquad \Longleftrightarrow \qquad \sigma(i_p)=i_q.
    \label{eq:restricted_permutation}
\end{equation}
and the superscripts denote the transformation of the function under permutation,
\begin{equation*}
    f^{\sigma}(g) = f(\sigma g)\qquad \forall g \in \Sm\; (\Sk).
\end{equation*}
When applied to a collection of node features, narrowing simply saves the features in nodes in \sm{\ssseq{i}{k}} and the average feature and then discards the rest.
More generally, if the activation is a multi-index tensor whose indices correspond to individual nodes, narrowing forms new tensors by averaging over the nodes \emph{not} in \sm{\{\ssseq{i}{k}\}}.

\vspace{\subsectionpadding}
\subsubsection{Promotion}
\vspace{\subsectionpadding}
Promotion is the opposite of narrowing in that it takes a function \m{g\colon \Sbb_k\to\RR^d}
and extends it to a function on \m{\Sm}.
We therefore apply the same construction as in Definition~\ref{defn:narrowing} in reverse.
\begin{definition}
    Let \m{(\sseq{j}{m})} be an ordered set of indices with an ordered subset \m{(\sseq{i}{k})}.
    Let $u$, $s$, $t$, $\acute{u}$, and $\grave{s}$ be as in Definition~\ref{defn:narrowing}.
    Given a function \m{g\colon\Sbb_k\to\RR^d}, 
we define the \df{promotion} of \m{g} to \m{\Sm} as the function:
    \begin{equation}
    g\up^{(\ssseq{j}{m})}(\tau)=
        \begin{cases}
            ~g(u) & \text{if there exist } \;u\tin \Sk,\; s\tin \Sbb_{m-k}\text{ ~such that~ } \tau = \acute{u} \grave{s} t, \\
           ~0 &\text{otherwise}. 
        \end{cases}
    \end{equation}
\end{definition}
In Section~4 of the supplement we show that any such $u$ and $s$ are unique
and consequently our definition is independent of the 
choice of \m{u} and \m{s}.
Narrowing is the pseudoinverse of promotion in the sense that for any \m{g\colon \Sk\to\RR^d}
\begin{equation*}
g\up^{(\ssseq{j}{m})}\dn_{(\ssseq{i}{k})}=g.
\end{equation*}
In contrast, 
narrowing followed by promotion is typically a lossy operation that does not preserve a function.
\changed{
    Similarly to narrowing, promotion obeys the equivariance property:
    \begin{equation*}
        (g^{\sigma'})\up^{(\ssseq{j}{m})}=(g\up^{(\ssseq{j}{m})})^{\sigma},
    \end{equation*} 
    where $\sigma$ and $\sigma'$ are defined as in~\eqref{eq:restricted_permutation}.
}
In the case of node features, promotion simply copies the node features into the new local graph.
For a multi-index tensors whose indices correspond to individual nodes, promotion zero-pads the tensor, adding indices for the new nodes.

\vspace{\subsectionpadding}
\subsection{Autobahn neurons}\label{ssec:auto_conv_neuron}
\vspace{\subsectionpadding}
Stated most generally, an Autobahn neuron $\neuron_j^\ell$ operates as follows.
Let \sm{\neuron^\ell_j} be a neuron whose 
local graph \sm{\rfd^\ell_j} is defined on the nodes \sm{\left\{v_{a_1},\ldots, v_{a_m}\right\}}.
Denote by 
\sm{f^{\ell-1}_{s_1},\ldots, f^{\ell-1}_{s_p}} 
the activations of the neurons in the previous layers whose local graphs overlap with \sm{\rfd^\ell_j}.
We denote the nodes in the local graph of the $z$'th overlapping neuron by 
\sm{(v_{a^z_1},\ldots, v_{a^z_{m_z}})} 
and define the intersections: 
\begin{equation*}
\cbrN{\sseq{b^z}{k_z}}=\cbrN{\sseq{a}{m}} \cap \cbrN{\sseq{a^z}{m_z}}. 
\end{equation*}
The operation performed by \sm{\neuron^\ell_j} in an Autobahn can then be summarized as  
follows:
\begin{compactenum}[~T1.]
\item Narrow each incoming activation \m{f^{\ell-1}_{s_z}} to the corresponding 
intersection to get \m{f^{\ell-1}_{s_z}\dn_{(\ssseq{b^z}{k_z})}}.   
\item Promote each of these to \m{(\sseq{a}{m})}\,: 
\vspace{-3pt}
\[ \tilde f_{s_z}=f^{\ell-1}_{s_z}\dn_{(\ssseq{b^z}{k_z})}\up^{(\ssseq{a}{m})}.
\]
Note that each \m{\tilde f_{s_z}} is \m{(\sseq{a}{m})}-permutation equivariant. 
\item Combine the results into a single function \m{\tilde f} by applying an aggregation function that is invariant to permutations of the set \sm{ \{ \sseq{\tilde f}{p}\}} within itself (for instance, averaging).
\item Apply one or more convolutions and nonlinearities over the local graph's automorphism group as described in Algorithm~\ref{algo:autoconv}.
\end{compactenum}

Sufficient conditions for the resulting network to obey global permutation equivariance are given below.
\begin{theorem}
    Let $\neuron_j^\ell$ be an Autobahn neuron in a neural network operating on a graph $\Gcal$.
    Let \sm{\rfd_j^\ell} be the local graph of \sm{\neuron_j^\ell} and denote \sm{\rfd_j^\ell}'s node set as \sm{\cbrN{v_{a_1},\ldots, v_{a_m}} \subset \cbrN{\sseq{v}{n}}} 
    and its edges as \m{\mathcal{E}_j^\ell = \left\{e_{kl}\right\}_{k,l\in \cbrN{v_{a_1},\ldots, v_{a_m}}}}.
    If the following three conditions hold then the resulting Autobahn obeys permutation equivariance.
\begin{compactenum}[~1.]
    \item For any permutation \m{\sigma\tin\Sn} applied to \m{\Gcal}, the resulting new network 
        \m{\Phi'} will have a neuron \sm{{\neuron'}^\ell_{j'}} with the same parameters that operates on a graph \sm{{\rfd_j^\ell}'}. The nodes of \sm{{\rfd_j^\ell}'} are \sm{\cbrN{v_{\sigma(a_1)},\ldots, v_{\sigma(a_m)}}} and its edges are  \sm{\cbrN{\,e_{\sigma(k)\sigma(l)}\:  \mid \: e_{kl}\in \mathcal{E}_j^\ell\,}}. 
    \item The output of the neuron is invariant with respect to all that permutations of $\Gcal$ that leave the nodes $\cbrN{v_{a_1},\ldots, v_{a_m}}$ in place.
    \item The output of the neuron is 
        equivariant to all permutations of the set $\cbrN{v_{a_1},\ldots, v_{a_m}}$ within itself.
\end{compactenum}
\end{theorem}
\vspace{-5pt}
A proof is given in Section~5 of the supplement.

\vspace{\subsectionpadding}
\subsection{Expressivity of Autobahn}
\vspace{\subsectionpadding}
To further understand the capabilities of Autobahn we will compare its theoretical expressivity to that of other graph neural networks.
First, we analyze Autobahn in the context of the $k$-Weisfeiler-Lehman ($k$-WL) algorithm and the $k$'th order network proposed in~\cite{maron2018invariant}.
Next, we compare Autobahn to the Graph Substructure Networks (GSN) proposed in~\cite{bouritsas2020improving}.

A common tool used to analyze the expressivity of message-passing neural networks is comparison against the $k$-WL algorithm \cite{cai1992optimal,geerts2020expressive}.
Here, information is repeatedly transferred between all possible ordered sets of $k$ nodes
and if the output differs between two graphs then they are not isomorphic.
It has since been determined that most message-passing neural networks are limited in their expressive power by the $2$-WL algorithm,
meaning certain graphs are fundamentally indistinguishable  by MPNNs \cite{xu2018powerful,morris2019weisfeiler}.
The WL algorithm is also closely related to the $k$'th order graph neural network from~\cite{maron2018invariant},
where activations correspond to all possible ordered sets of $k$ nodes 
and information is transferred between sets using tensor expansions and contractions.
This process corresponds to convolutions over all permutations of the graph's nodes \cite{thiede2020general}.
Here, a $k$'th order network has the expressive power of the $k$-WL.
However, the size of the activations grows combinatorially as we increase $k$, 
making the network infeasible for all but small values of $k$.

\begin{figure}
  \begin{center}
      \includegraphics[width=0.6\textwidth]{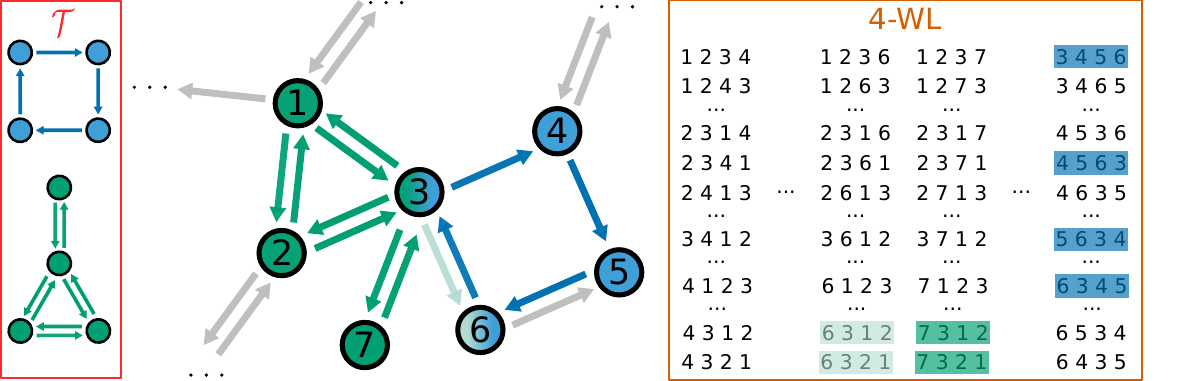}
  \end{center}
  \caption{
      Visual comparison of the activations in Autobahn with the data structures used in the $k$-WL algorithm and in a $k$'th order graph neural network.
      Whereas the $k$-WL and $k$'th order GNN operate over all possible ordered sequences of $k$ nodes (here $k=4$), Autobahn 
      specifically targets sequences believed to be important by identifying them through the isomorphism class of the corresponding subgraph.
      Here, one subgraph is isomorphic to the blue template and two are isomorphic to the green.
      Sequences corresponding to their automorphism groups are highlighted in the corresponding color.
      Using subsets of sequences corresponding to specific substructures allows Autobahn to perform higher-order computation without the combinatorial explosion in cost.
       }
  \vspace{1.0\figpadding}
  \label{fig:k_wl_flat}
\end{figure}
Autobahn, in contrast, constructs activations on specific subgraphs, corresponding to specifically chosen ordered sets of nodes.
As depicted in Figure~\ref{fig:k_wl_flat}, this makes the Autobahn activation sparse in the set of all possible node sets.
If these sets are chosen well, the Autobahn network can hopefully leverage the power of higher-order computation
without incurring a combinatorial increase in cost.
In Subsection~6.1 of the Supplement
we formalize this connection by showing that 
the operations in Autobahn can be performed "densely" using the $k$-th order network,
and that choosing an Autobahn template that covers all sets of $k$ nodes recovers a $k$-th order network.

The strategy of using subgraphs to improve the expressiveness of a 
neural network is shared by the GSN network \cite{bouritsas2020improving},
which augments a message-passing neural network with node features that count the isomorphism classes of subgraphs a particular node is in.
Consequently, it is reasonable to ask if using subgraphs for computation (as Autobahn does) 
gives any advantages compared to merely using them to create initial features.
In Subsection~6.2 of the Supplement
, we answer this in the affirmative.
Moreover, while it was noted that a GSN network would be able to reconstruct a graph from its $n\<-1$ subgraphs
if the reconstruction conjecture \cite{kelly1957congruence,ulam1960collection} held, it is not clear if this task could be accomplished if the reconstruction conjecture was false.
In contrast, we show that the high-order activations and the transfer of information using narrowing and promotion
allows Autobahn to reconstruct a graph using neurons operating on its subgraphs of size $n\<-1$
independently of the reconstruction conjecture.

\vspace{\sectionpadding}
\section{Molecular graphs on the Autobahn}\label{sec:experiments}
\vspace{\sectionpadding}
With the Autobahn formalism defined, we return to our motivating task of learning the properties of molecular graphs.
From the structure of organic molecules, we see that they often 
have a combination of a sparse chain-like ``backbone''
and cyclic structures such as aromatic rings.
The importance of these structures is further justified by the theory of molecular resonance.
Whereas in molecular graphs edges correspond to individual pairs of electrons, 
real electrons cannot be completely localized to single bonds.
To re-inject this physics into graph representations of molecules, chemists construct ``resonance structures'':
alternate molecular graphs formed by concertedly moving the electrons in a molecular graph.
Importantly, the rules of chemical valency ensure that these motions occur almost exclusively 
on paths or cycles within the graph.
In fact, cycle and path featurizations have already been used successfully in cheminformatic applications \cite{dixon2006phase}.

Motivated by these considerations, we will choose our local graphs to correspond to directed cycles and paths in graph.
This has the additional advantage that the one-dimensional convolutions given by~\eqref{eq:one_dimensional_conv}
are equivariant to the graph's automorphism group, and can be used directly.
\begin{figure}
    \centering
    \begin{subfigure}[]{0.75\columnwidth}
        \includegraphics[width=.98\columnwidth]{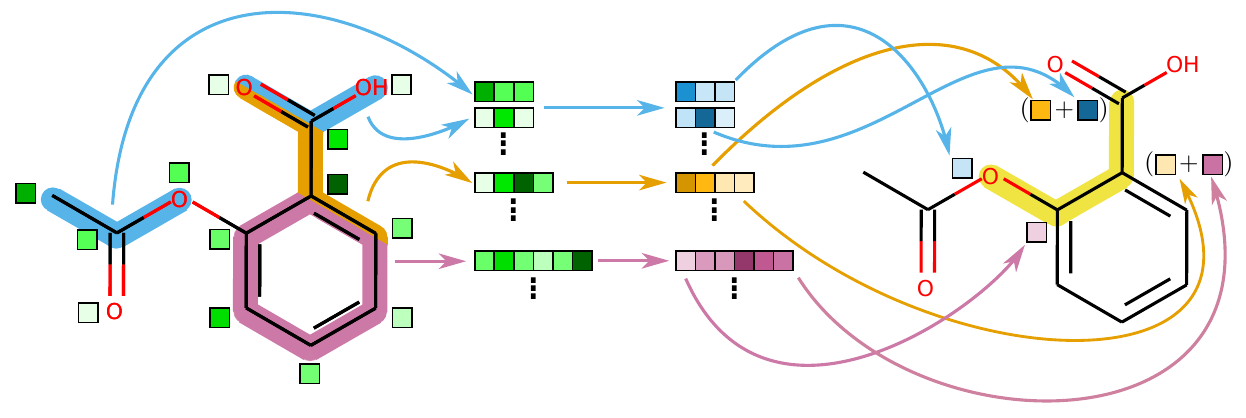}
    \end{subfigure}
    \begin{subfigure}[]{0.2\columnwidth}
{\footnotesize
        \begin{tikzpicture}[node distance=6pt,blocknode/.style={rectangle,draw=black,rounded corners},>=latex]
        \node (input) {};
        \node (identity) [below=of input] {};
        \node[blocknode] (conv1) [below=of identity] {Path / Cycle Conv};
        \node[blocknode] (conv2) [below=20 pt of conv1] {Path / Cycle Conv};
        \node[circle,draw=black] (resid) [below=of conv2] {+};
        \node (output) [below=15 pt of resid] {};
        \node (conv1right) [right=of conv1] {};
        \node (conv2right) [right=of conv2] {};
        \draw[->] (input) -- (conv1);
        \draw[->] (conv1) -- (conv2) node[midway, right] {{\footnotesize ReLU}};
        \draw[->] (conv2) -- (resid);
        \draw[->] (resid) -- (output) node[midway, right] {{\footnotesize ReLU}};
        \draw[->] (identity) to [out=0,in=90] (conv1right.mid) -- (conv2right.mid) to [out=270,in=0] (resid);
    \end{tikzpicture}
}
    \end{subfigure}
    \vspace{-10pt}
    \caption{
        The internal structure of a single layer in the Autobahn architecture.
        We extract all path and cycle subgraphs of fixed length 
        and their corresponding activations.
        (For compactness and readability, only some activations are shown.)
        We then apply a series of convolutional layers.  A block diagram for this step is given on the right; each isomorphism class of reference domains has its own weights.
        Finally, we construct the activations for the next layer by narrowing and promoting between subgraphs
        and summing over the resulting promoted activations.
    \label{fig:block-diagram}}
\end{figure}
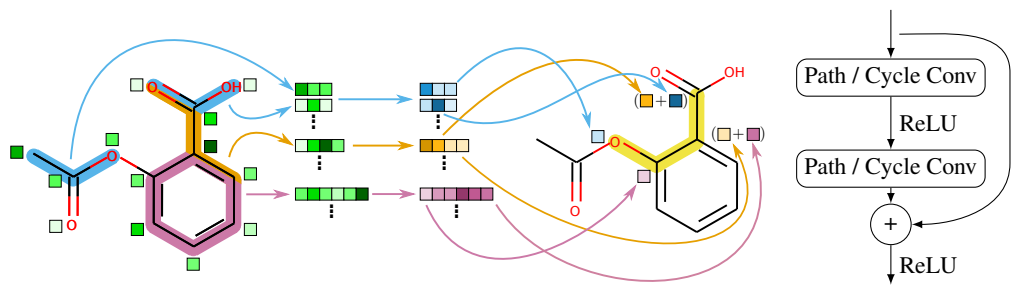
To construct the neurons for our architecture,
we extract all paths of length three through six in the graph
and  all cycles of five or six elements.
These cycle lengths were chosen because cycles of five and six elements are particularly common 
in chemical graphs.
For each path or cycle, we construct two neurons corresponding to two ways of traversing the graph:
for cycles, this corresponds to clockwise or anticlockwise rotation of the cycle, and for paths this
sets one of the ends to be the ``initial'' node.

\sisetup{separate-uncertainty=true}
\begin{table}[t]
    \centering  
    {\small
     \begin{tabular}{lccccc}
    \toprule
    Model & ZINC 10k & ZINC full & MolPCBA  & MolHIV & MUV \\ 
     & (MAE, $\dn$) & (MAE, $\dn$) & (AP $\up$) & (ROCAUC $\up$) & (AP $\up$) \\ 
    \midrule
    GCN    & \num{0.367 \pm 0.011} & N/A  & \num{0.222 \pm 0.002} & \num{0.788 \pm 0.080}  & N/A          \\
    GSN & \num{0.108 \pm 0.018}   & N/A  & N/A  & \num{0.780 \pm 0.010}  &   N/A          \\
    DGN & \num{0.169 \pm 0.003}   & N/A   &  N/A          & {\bfseries \num[detect-all=true]{0.797 \pm 0.010}}  &   N/A          \\
    GINE-E & \num{0.252 \pm 0.015} & \num{0.088 \pm 0.002} & \num{0.227 \pm 0.003} & \num{0.788 \pm 0.080}  & \makebox[\widthof{\num{0.091 \pm 0.000}}][l]{0.091} \\
    HIMP & \num{0.151 \pm 0.006}   & \num{0.032 \pm 0.002} &  {\bfseries \num[detect-all=true]{0.274 \pm 0.003}} & \num{0.788 \pm 0.08} & \num{0.114 \pm 0.041} \\
    \textbf{Ours} & {\bfseries \num[detect-all=true]{0.106 \pm 0.004}}   & {\bfseries \num[detect-all=true]{0.029 \pm 0.001}} &   \makebox[\widthof{\num{0.270 \pm 0.000}}][l]{\num{0.270}}            & \num{0.780 \pm 0.003} & {\bfseries \num[detect-all=true]{0.119 \pm 0.005}}  \\
    \end{tabular}
    }\vspace{3pt}
    \caption{Performance of our Autobahn architecture on two splits of the ZINC dataset and three datasets in the OGB benchmark family, compared with other recent message passing architectures.  ZINC experiments use MAE (lower is better); for all other metrics higher is better. Baselines  were taken from~\cite{fey2020hierarchical, brossard2020graph, beaini2021directional} and~\cite{bouritsas2020improving}.
    \vspace{1.5\figpadding}
    }
\end{table}
We then construct initial features for each neuron by embedding atom and bond identities as
categorical variables. 
Embedded atom identities are then directly assigned to the corresponding points in each path or cycle.
To assign the embedded bond identities, we arbitrarily assign each bond to the preceding node in the 
traversals mentioned above.  
Since we construct a neuron for both traversal directions, this procedure does not break permutation equivariance 
of the architecture.
Following the initial featurization, we then construct layers using the four step procedure
described in Section~\ref{sec:autobahn}.
The layer is illustrated in Figure~\ref{fig:block-diagram}.

Full details of the model, including training hyper-parameters and architecture details,
are available in the Supplement;
code is freely available at \url{https://github.com/risilab/Autobahn}.
We present empirical results from an implementation of our architecture on two subsets of the ZINC dataset using the data splits from~\cite{fey2020hierarchical}, 
as well as three standardized tasks from Open Graph Benchmark.
All datasets are released under the MIT license.
Baselines were taken from~\cite{fey2020hierarchical, brossard2020graph, beaini2021directional, bouritsas2020improving}.
Our automorphism-based neural network achieves results competitive with modern MPNNs.

\vspace{\sectionpadding}
\section{Conclusion}\label{sec:conclusion}
\vspace{\sectionpadding}
In this paper, we have introduced Automorphism-based Neural Networks (Autobahn), a new framework for constructing neural networks on graphs.
To build an Autobahn, we first choose a collection of template graphs.
We then break our input graph into a collection of local graphs, each isomorphic to a template.
Computation proceeds on each local graph by applying convolutions equivariant to the template's automorphism group,
and by transferring information between the local graphs using two operators we refer to as ``narrowing'' and ``promotion''.
MPNNs are specific examples of Autobahn networks constructed by choosing star-shaped templates applied to local neighborhoods.
Similarly, applying Autobahn to a grid graph recovers steerable CNNs.
Our experimental results show that Autobahn networks can be competitive with modern MPNNs on several molecular learning tasks.

We expect the choice of substructure to critically influence Autobahn performance.
In future work we hope to explore the space of new models opened up by our theory.
In learning situations where much is known about the graphs' structure,
we believe practitioners will be able to choose templates that correspond to desired inductive biases, 
giving improved results.
For instance, in future work we hope to improve our results on molecular graphs by adding templates that correspond to specific functional groups.
For arbitrary graphs, it is not clear that the star graphs used by MPNNs are optimal or just a historical accident.
It is possible that  other ``generic'' templates exist that give reasonable results for a wide variety of graphs.
For example,~\cite{perozzi2014deepwalk,toenshoff2021graph} used path activations  in conjunction with stochastic sampling strategies and the architecture in~\cite{flam2021neural} can be viewed as using tree-like substructures. 
By further exploring the space of possible templates, we hope to construct richer graph neural networks that more naturally reflect the graphs on which they operate.

\vspace{\subsectionpadding}
\subsection{Broader Impacts}
\vspace{\subsectionpadding}
In our framework, the choice of template reflects practitioners' beliefs about which graph substructures are important for determining its properties.
For social networks
communities from different cultural backgrounds might result in graphs with differing topologies.
Consequently, when applying Autobahn to these graphs care must be taken that chosen templates do not implicitly bias our networks towards specific cultural understandings.

\section{Acknowledgements}
This project was supported by DARPA “Physics of AI” grant number HR0011837139, and used computational resources acquired through NSF MRI 1828629. The Flatiron Institute is a division of the Simons Foundation. We thank Sonya Hanson, John Herr, Joe Paggi, and Helen Yu for useful feedback.

\clearpage

\bibliography{gnn}

\begin{thebibliography}{10}

\bibitem{scarselli2008graph}
Franco Scarselli, Marco Gori, Ah~Chung Tsoi, Markus Hagenbuchner, and Gabriele
  Monfardini.
\newblock The graph neural network model.
\newblock {\em IEEE Transactions on Neural Networks}, 20(1):61--80, 2008.

\bibitem{perozzi2014deepwalk}
Bryan Perozzi, Rami Al-Rfou, and Steven Skiena.
\newblock Deepwalk: Online learning of social representations.
\newblock In {\em Proceedings of the 20th ACM SIGKDD International Conference
  on Knowledge Discovery and Data Mining}, pages 701--710, 2014.

\bibitem{HenaffLeCun2015}
M.~Henaff, J.~Bruna, and Y.~LeCun.
\newblock Deep convolutional networks on graph-structured data.
\newblock {\em arXiv preprint arXiv:1506.05163}, 06 2015.

\bibitem{Defferrard2016}
Micha\"el Defferrard, Xavier Bresson, and Pierre Vandergheynst.
\newblock Convolutional neural networks on graphs with fast localized spectral
  filtering.
\newblock In {\em Advances in Neural Information Processing Systems (NeurIPS)},
  2016.

\bibitem{bronstein2017geometric}
Michael~M Bronstein, Joan Bruna, Yann LeCun, Arthur Szlam, and Pierre
  Vandergheynst.
\newblock Geometric deep learning: going beyond {E}uclidean data.
\newblock {\em IEEE Signal Processing Magazine}, 34(4):18--42, 2017.

\bibitem{KipfWelling2017}
T.~N. Kipf and M.~Welling.
\newblock Semi-supervised classification with graph convolutional networks.
\newblock In {\em International Conference on Learning Representations (ICLR)},
  2017.

\bibitem{Gilmer2017}
Justin Gilmer, Samuel~S. Schoenholz, Patrick~F. Riley, Oriol Vinyals, and
  George~E. Dahl.
\newblock Neural message passing for quantum chemistry.
\newblock In {\em Proceedings of International Conference on Machine Learning
  (ICML)}, 2017.

\bibitem{xu2018powerful}
Keyulu Xu, Weihua Hu, Jure Leskovec, and Stefanie Jegelka.
\newblock How powerful are graph neural networks?
\newblock In {\em International Conference on Learning Representations (ICLR)},
  2019.

\bibitem{hu2019strategies}
Weihua Hu, Bowen Liu, Joseph Gomes, Marinka Zitnik, Percy Liang, Vijay~S.
  Pande, and Jure Leskovec.
\newblock Strategies for pre-training graph neural networks.
\newblock In {\em International Conference on Learning Representations (ICLR)},
  2020.

\bibitem{li2019deepgcns}
Guohao Li, Matthias M{\"{u}}ller, Ali~K. Thabet, and Bernard Ghanem.
\newblock Deep{GCN}s: Can {GCN}s go as deep as {CNN}s?
\newblock In {\em {IEEE/CVF} International Conference on Computer Vision,
  {ICCV}}, 2019.

\bibitem{arvind2020weisfeiler}
Vikraman Arvind, Frank Fuhlbr{\"u}ck, Johannes K{\"o}bler, and Oleg Verbitsky.
\newblock On {W}eisfeiler-{L}eman invariance: Subgraph counts and related graph
  properties.
\newblock {\em Journal of Computer and System Sciences}, 113:42--59, 2020.

\bibitem{chen2020can}
Zhengdao Chen, Lei Chen, Soledad Villar, and Joan Bruna.
\newblock Can graph neural networks count substructures?
\newblock In {\em Advances in Neural Information Processing Systems (NeurIPS)},
  2020.

\bibitem{garg2020generalization}
Vikas Garg, Stefanie Jegelka, and Tommi Jaakkola.
\newblock Generalization and representational limits of graph neural networks.
\newblock In {\em icml}, pages 3419--3430. PMLR, 2020.

\bibitem{korotcov2017comparison}
Alexandru Korotcov, Valery Tkachenko, Daniel~P Russo, and Sean Ekins.
\newblock Comparison of deep learning with multiple machine learning methods
  and metrics using diverse drug discovery data sets.
\newblock {\em Molecular {P}harmaceutics}, 14(12):4462--4475, 2017.

\bibitem{bock2019review}
Frederic~E Bock, Roland~C Aydin, Christian~J Cyron, Norbert Huber, Surya~R
  Kalidindi, and Benjamin Klusemann.
\newblock A review of the application of machine learning and data mining
  approaches in continuum materials mechanics.
\newblock {\em Frontiers in Materials}, 6:110, 2019.

\bibitem{haghighatlari2019advances}
Mojtaba Haghighatlari and Johannes Hachmann.
\newblock Advances of machine learning in molecular modeling and simulation.
\newblock {\em Current Opinion in Chemical Engineering}, 23:51--57, 2019.

\bibitem{pollice2021data}
Robert Pollice, Gabriel dos Passos~Gomes, Matteo Aldeghi, Riley~J Hickman,
  Mario Krenn, Cyrille Lavigne, Michael Lindner-D’Addario, AkshatKumar Nigam,
  Cher~Tian Ser, Zhenpeng Yao, et~al.
\newblock Data-driven strategies for accelerated materials design.
\newblock {\em Accounts of Chemical Research}, 54(4):849--860, 2021.

\bibitem{Duvenaud2015}
David Duvenaud, Dougal Maclaurin, Jorge Aguilera-Iparraguirre, Rafael
  Gomez-Bombarelli, Timothy Hirzel, Alan Aspuru-Guzik, and Ryan~P. Adams.
\newblock Convolutional networks on graphs for learning molecular fingerprints.
\newblock In {\em Advances in Neural Information Processing Systems (NeurIPS)},
  2015.

\bibitem{hamilton2017inductive}
William~L Hamilton, Rex Ying, and Jure Leskovec.
\newblock Inductive representation learning on large graphs.
\newblock In {\em Advances in Neural Information Processing Systems (NeurIPS)},
  pages 1025--1035, 2017.

\bibitem{alsentzer2020subgraph}
Emily Alsentzer, Samuel Finlayson, Michelle Li, and Marinka Zitnik.
\newblock Subgraph neural networks.
\newblock In {\em Advances in Neural Information Processing Systems (NeurIPS)},
  volume~33, 2020.

\bibitem{bouritsas2020improving}
Giorgos Bouritsas, Fabrizio Frasca, Stefanos Zafeiriou, and Michael~M
  Bronstein.
\newblock Improving graph neural network expressivity via subgraph isomorphism
  counting.
\newblock {\em arXiv preprint arXiv:2006.09252}, 2020.

\bibitem{JunctionTreeVAE_ICML2018}
Wengong Jin, Regina Barzilay, and Tommi Jaakkola.
\newblock Junction tree variational autoencoder for molecular graph generation.
\newblock In {\em Proceedings of International Conference on Machine Learning
  (ICML)}, 2018.

\bibitem{fey2020hierarchical}
Matthias Fey, Jan-Gin Yuen, and Frank Weichert.
\newblock Hierarchical inter-message passing for learning on molecular graphs.
\newblock In {\em Graph Representation Learning and Beyond (GRL+) Workshop at
  ICML 2020}, 2020.

\bibitem{KondorTrivedi2018}
Risi Kondor and Shubhendu Trivedi.
\newblock On the generalization of equivariance and convolution in neural
  networks to the action of compact groups.
\newblock In {\em Proceedings of International Conference on Machine Learning
  (ICML)}, 2018.

\bibitem{Cohen2016}
Taco~S. Cohen and Max Welling.
\newblock {Group equivariant convolutional networks}.
\newblock In {\em Proceedings of International Conference on Machine Learning
  (ICML)}, 2016.

\bibitem{Cohen2017}
Taco~S. Cohen and Max Welling.
\newblock Steerable {CNN}s.
\newblock In {\em International Conference on Learning Representations (ICLR)},
  2017.

\bibitem{CohenCNNhomo}
Taco~S. Cohen, Mario Geiger, and Maurice Weiler.
\newblock A general theory of equivariant {CNN}s on homogeneous spaces.
\newblock In {\em Advances in Neural Information Processing Systems (NeurIPS)},
  2019.

\bibitem{DeepSets}
Manzil Zaheer, Satwik Kottur, Siamak Ravanbakhsh, Barnabas P{\'o}czos, Russ~R
  Salakhutdinov, and Alexander~J Smola.
\newblock Deep sets.
\newblock In {\em Advances in Neural Information Processing Systems (NeurIPS)},
  2017.

\bibitem{thiede2020general}
Erik~H. Thiede, Truong~Son Hy, and Risi Kondor.
\newblock The general theory of permutation equivariant neural networks and
  higher order graph variational encoders.
\newblock {\em arXiv preprint arXiv:2004.03990}, 2020.

\bibitem{maron2018invariant}
Haggai Maron, Heli Ben-Hamu, Nadav Shamir, and Yaron Lipman.
\newblock Invariant and equivariant graph networks.
\newblock In {\em International Conference on Learning Representations (ICLR)},
  2018.

\bibitem{maron2019universality}
Haggai Maron, Ethan Fetaya, Nimrod Segol, and Yaron Lipman.
\newblock On the universality of invariant networks.
\newblock In {\em Proceedings of International Conference on Machine Learning
  (ICML)}, 2019.

\bibitem{maron2019provably}
Haggai Maron, Heli Ben-Hamu, Hadar Serviansky, and Yaron Lipman.
\newblock Provably powerful graph networks.
\newblock In {\em Advances in Neural Information Processing Systems (NeurIPS)},
  2019.

\bibitem{kondor2018covariant}
Truong~Son Hy, Shubhendu Trivedi, Horace Pan, Brandon~M. Anderson, and Risi
  Kondor.
\newblock Predicting molecular properties with covariant compositional
  networks.
\newblock {\em The Journal of Chemical Physics}, 148(24):241745, 2018.

\bibitem{dehaan2020natural}
Pim de~Haan, Taco~S. Cohen, and Max Welling.
\newblock Natural graph networks.
\newblock In {\em Advances in Neural Information Processing Systems (NeurIPS)},
  2020.

\bibitem{cordella2001improved}
Luigi~Pietro Cordella, Pasquale Foggia, Carlo Sansone, and Mario Vento.
\newblock An improved algorithm for matching large graphs.
\newblock In {\em 3rd IAPR-TC15 Workshop on Graph-based Representations in
  Pattern Recognition}. Citeseer, 2001.

\bibitem{cordella2004sub}
Luigi~P Cordella, Pasquale Foggia, Carlo Sansone, and Mario Vento.
\newblock A (sub)graph isomorphism algorithm for matching large graphs.
\newblock {\em IEEE Transactions on Pattern Analysis and Machine Intelligence},
  26(10):1367--1372, 2004.

\bibitem{junttila2007engineering}
Tommi Junttila and Petteri Kaski.
\newblock Engineering an efficient canonical labeling tool for large and sparse
  graphs.
\newblock In {\em 2007 Proceedings of the Ninth Workshop on Algorithm
  Engineering and Experiments ({ALENEX})}. SIAM, 2007.

\bibitem{han2013turboiso}
Wook-Shin Han, Jinsoo Lee, and Jeong-Hoon Lee.
\newblock Turboiso: Towards ultrafast and robust subgraph isomorphism search in
  large graph databases.
\newblock In {\em Proceedings of the 2013 ACM SIGMOD International Conference
  on Management of Data}, 2013.

\bibitem{morris2019weisfeiler}
Christopher Morris, Martin Ritzert, Matthias Fey, William~L Hamilton, Jan~Eric
  Lenssen, Gaurav Rattan, and Martin Grohe.
\newblock Weisfeiler and {L}eman go neural: Higher-order graph neural networks.
\newblock In {\em Proceedings of the AAAI Conference on Artificial
  Intelligence}, 2019.

\bibitem{vignac2020building}
Clement Vignac, Andreas Loukas, and Pascal Frossard.
\newblock Building powerful and equivariant graph neural networks with
  structural message-passing.
\newblock {\em arXiv preprint arXiv:2006.15107}, 2020.

\bibitem{finzi2021practical}
Marc Finzi, Max Welling, and Andrew~Gordon Wilson.
\newblock A practical method for constructing equivariant multilayer
  perceptrons for arbitrary matrix groups.
\newblock {\em arXiv preprint arXiv:2104.09459}, 2021.

\bibitem{cai1992optimal}
Jin-Yi Cai, Martin F{\"u}rer, and Neil Immerman.
\newblock An optimal lower bound on the number of variables for graph
  identification.
\newblock {\em Combinatorica}, 12(4):389--410, 1992.

\bibitem{geerts2020expressive}
Floris Geerts.
\newblock The expressive power of kth-order invariant graph networks.
\newblock {\em arXiv preprint arXiv:2007.12035}, 2020.

\bibitem{kelly1957congruence}
Paul~J Kelly.
\newblock A congruence theorem for trees.
\newblock {\em Pacific Journal of Mathematics}, 7(1):961--968, 1957.

\bibitem{ulam1960collection}
Stanislaw~M Ulam.
\newblock {\em A collection of mathematical problems}.
\newblock Interscience Publishers, 1960.

\bibitem{dixon2006phase}
Steven~L Dixon, Alexander~M Smondyrev, Eric~H Knoll, Shashidhar~N Rao, David~E
  Shaw, and Richard~A Friesner.
\newblock {PHASE}: a new engine for pharmacophore perception, {3D QSAR} model
  development, and {3D} database screening: 1. {M}ethodology and preliminary
  results.
\newblock {\em Journal of Computer-Aided Molecular Design}, 20(10):647--671,
  2006.

\bibitem{brossard2020graph}
R{\'e}my Brossard, Oriel Frigo, and David Dehaene.
\newblock Graph convolutions that can finally model local structure.
\newblock {\em arXiv preprint arXiv:2011.15069}, 2020.

\bibitem{beaini2021directional}
Dominique Beaini, Saro Passaro, Vincent L{\'e}tourneau, William~L Hamilton,
  Gabriele Corso, and Pietro Li{\`o}.
\newblock Directional graph networks.
\newblock In {\em ICLR 2021 Workshop on Geometrical and Topological
  Representation Learning}, 2021.

\bibitem{toenshoff2021graph}
Jan Toenshoff, Martin Ritzert, Hinrikus Wolf, and Martin Grohe.
\newblock Graph learning with 1{D} convolutions on random walks.
\newblock {\em arXiv preprint arXiv:2102.08786}, 2021.

\bibitem{flam2021neural}
Daniel Flam-Shepherd, Tony~C Wu, Pascal Friederich, and Alan Aspuru-Guzik.
\newblock Neural message passing on high order paths.
\newblock {\em Machine Learning: Science and Technology}, 2021.

\end{thebibliography}
\bibliographystyle{unsrt}

\end{document}


\maketitle

\section{Activations as functions on a group}\label{sec:group_activations}
In the Autobahn formalism, we make extensive use of the fact 
that the activations of a group-equivariant neural network
can be treated as functions on the same group.
Here we give a brief review for the unfamiliar reader.
This formalism is also covered in detail in Sections~3 and~4 of Reference~\cite{KondorTrivedi2018},
although under slightly different conventions.

Consider a space $\mathcal{X}$ acted on by a group $G$:
at every point $x$ in $\mathcal{X}$, we can apply a group element
$g \in G$, which maps $x$ to another point in $\mathcal{X}$.
The action of the group on $\mathcal{X}$ induces an action on functions of $\mathcal{X}$.
We define an operator $T_g$ acting on functions $f :\mathcal{X}\to \mathbb{C}$ as follows.
\footnote{Note this is a different convention for the action of group elements on functions from the one described in Reference~\cite{KondorTrivedi2018}. 
The choice of whether to use $T_g$ operator described here or the $\mathbb{T}_g$ operator described in the reference is a matter of personal preference as they are inverses.}
\begin{equation}
    T_g \left( f\right)\left( x \right) = f\left( g\left( x \right) \right).
    \label{eq:defn_of_group_action}
\end{equation}
The inputs to group-equivariant neural networks are precisely 
functions on such spaces.
For instance, for standard convolutional layers acting on images, 
each point on the space is a single pixel and the group of translation moves between pixels.
The RGB value of each pixel is then vector-valued function of $\mathcal{X}$.

But while the input is a function on $\mathcal{X}$, representations internal to the network
can be more general.
For instance, consider a neural network that is given a list of objects and attempts to learn an adjacency matrix.
Here, $\mathcal{X}$ is the list of objects, and the object identity is a function on the list.
But the output space is a function on \textit{pairs} of objects: a different space.

Fortunately, the complexity of dealing with a myriad of spaces can be avoided by mapping 
functions from their individual spaces to functions on $G$.
This allows all possible spaces and group actions to be treated using a single (albeit abstract) formalism,
simplifying definitions and proofs.
For simplicity, we will assume that $G$ is transitive on $\mathcal{X}$: 
for all $x,y \in \mathcal{X}$, there exists a group element $g$
such that $g(x) = y$.
(If the $G$ is not transitive, we can simply apply this procedure on every orbit of $G$ and concatenate the results along a channel dimension.)
The construction proceeds as follows.  
We arbitrarily choose an initial point $x_0$ in $\mathcal{X}$ to act as the origin.
Then, we construct the function
\begin{equation}
    f_G \left( \sigma \right) = f\left( \sigma \left( x_0 \right) \right)\quad \forall \sigma \in G.
\end{equation}
Since we have assumed that $G$ is transitive, this is an injective map into functions on $G$ that preserves all of the information in $f$.
\begin{figure}[h]
    \centerline{\includegraphics[width=0.7\textwidth]{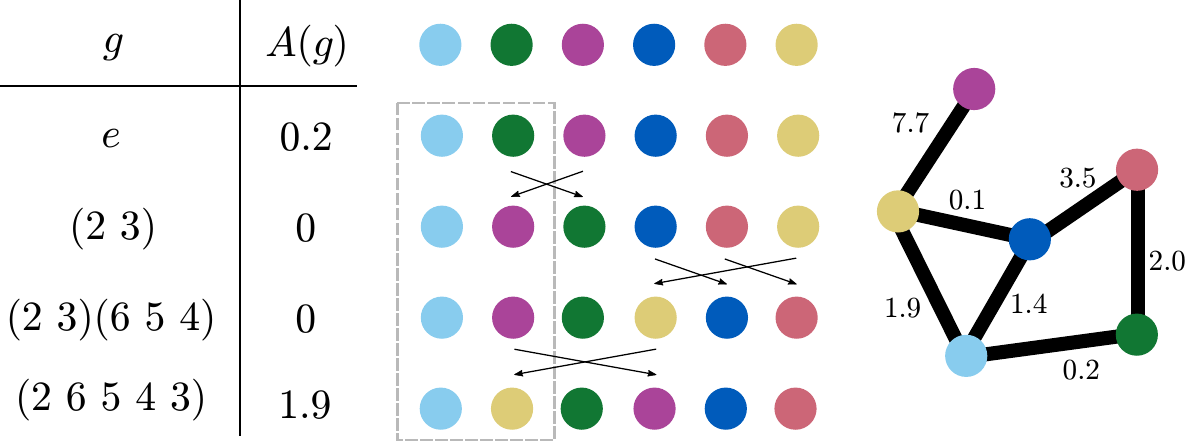}}
    \caption{
        Example showing how a function on the symmetric group can be constructed from
        neural network activations on the edges of a graph.
        To construct the function, we first 
        list the vertices in arbitrary order.
        Then, for every permutation, we permute the indices and 
        look at two arbitrarily chosen vertices (here we have chosen the first two).
        If these vertices form an edge, the function takes the edge activation as its value.
        Otherwise, the function takes a value of zero.
    }
    \label{fig:coset_construction}
\end{figure}
In Figure~\ref{fig:coset_construction}, we depict this procedure, 
mapping the adjacency matrix for a graph with size 6 vertices to the $S_6$,
the group of all permutations of six elements.
Subsequent group actions then transform the resulting function as follows.
\begin{equation}
    T_g f_G \left( \sigma \right) = f \left( \sigma  (g x_0) \right) = f_G \left( \sigma g \right). 
\end{equation}

\section{Equivariance of Automorphism-based Neurons}\label{sec:neuron_equivariance}

Here, we prove that the Algorithm~1 is equivariant to permutation.
\begin{theorem}
    Let $\Gcal$ be a graph of $n$ vertices and let $\sigma\in\Sn$.
The neuron $\neuron^\ell$ described in Algorithm~1 obeys
\begin{equation}
    \neuron^\ell \left( T_\sigma f^{\ell-1}\right) = T_\sigma'\neuron^{\ell} \left( f^{\ell-1}\right)
\end{equation}
\end{theorem}
\begin{proof}
Denote the permutation of $\Gcal$ by $\sigma$ by $\bar{\Gcal}$.
Applying $\neuron^\ell$  to $\bar{\Gcal}$ constructs a matching $\bar{\mu}$.
Since matching is accomplished up to an element in $\Tcal$'s automorphism group,
there exists $v \in \Aut ( \Tcal)$ such that 
\begin{equation*}
    v \mu = \bar{\mu} \sigma
\end{equation*}
Denoting convolution over $\Aut ( \Tcal )$ as $\ast$, we have 
\begin{align*}
    \neuron^\ell \left( T_\sigma f^{\ell-1}\right) &= {T'}_{{\bar{\mu}}^{-1}} \left(\nu \left( \left( {T}_{{\bar{\mu}}} T_\sigma f^{\ell-1} \right) \ast w + b \right) \right) \\
    &= {T'}_{{\bar{\mu}}^{-1}} \left(\nu \left( \left( {T}_{v} T_\mu f^{\ell-1} \right) \ast w + b \right) \right) \\
    &= {T'}_{{\bar{\mu}}^{-1}} {T'}_{v} \left(\nu \left( \left( T_\mu f^{\ell-1} \right) \ast w + b \right) \right) \\
    &= {T'}_{{\sigma}^{-1}} {T'}_{\mu^{-1}} \left(\nu \left( \left( T_\mu f^{\ell-1} \right) \ast w + b \right) \right).
\end{align*}
which proves equivariance.  
Note the third line follows from equivariance of convolution over $\Aut ( \Tcal )$ and the fact that $b$ is invariant to elements in $\Aut ( \Tcal )$.
\end{proof}

\section{Application of Autobahn to grid graphs}

Here, we discuss the application of Autobahn to grid graphs and show how the ideas in Autobahn can be used to recover 
the standard convolutional and steerable CNN (p4m) architectures
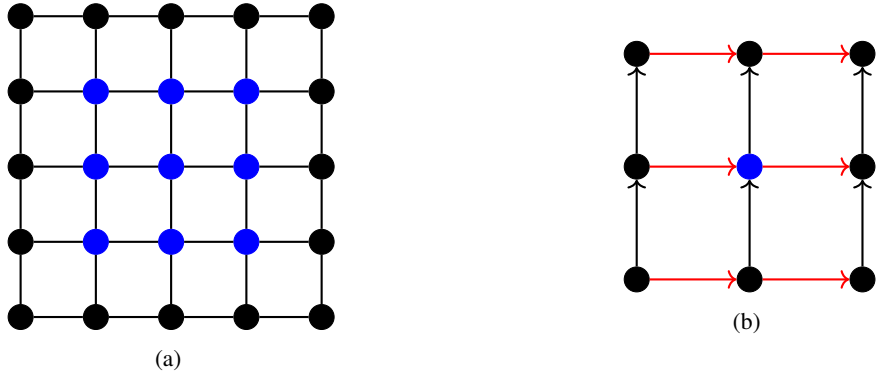
\begin{figure}
    \begin{subfigure}[]{0.5\columnwidth}
        \centering
\begin{tikzpicture}[scale=1.0, every node/.style={transform shape}]
    \node[shape=circle,draw=black, fill=black] (AA) at (0,0) {};
    \node[shape=circle,draw=black, fill=black] (BA) at (1,0) {};
    \node[shape=circle,draw=black, fill=black] (CA) at (2,0) {};
    \node[shape=circle,draw=black, fill=black] (DA) at (3,0) {};
    \node[shape=circle,draw=black, fill=black] (EA) at (4,0) {};
    \node[shape=circle,draw=black, fill=black] (AB) at (0,1) {};
    \node[shape=circle,draw=blue, fill=blue] (BB) at (1,1) {};
    \node[shape=circle,draw=blue, fill=blue] (CB) at (2,1) {};
    \node[shape=circle,draw=blue, fill=blue] (DB) at (3,1) {};
    \node[shape=circle,draw=black, fill=black] (EB) at (4,1) {};
    \node[shape=circle,draw=black, fill=black] (AC) at (0,2) {};
    \node[shape=circle,draw=blue, fill=blue] (BC) at (1,2) {};
    \node[shape=circle,draw=blue, fill=blue] (CC) at (2,2) {};
    \node[shape=circle,draw=blue, fill=blue] (DC) at (3,2) {};
    \node[shape=circle,draw=black, fill=black] (EC) at (4,2) {};
    \node[shape=circle,draw=black, fill=black] (AD) at (0,3) {};
    \node[shape=circle,draw=blue, fill=blue] (BD) at (1,3) {};
    \node[shape=circle,draw=blue, fill=blue] (CD) at (2,3) {};
    \node[shape=circle,draw=blue, fill=blue] (DD) at (3,3) {};
    \node[shape=circle,draw=black, fill=black] (ED) at (4,3) {};
    \node[shape=circle,draw=black, fill=black] (AE) at (0,4) {};
    \node[shape=circle,draw=black, fill=black] (BE) at (1,4) {};
    \node[shape=circle,draw=black, fill=black] (CE) at (2,4) {};
    \node[shape=circle,draw=black, fill=black] (DE) at (3,4) {};
    \node[shape=circle,draw=black, fill=black] (EE) at (4,4) {};
    
    \path [-, thick] (AA) edge node[left] {} (BA);
    \path [-, thick] (BA) edge node[left] {} (CA);
    \path [-, thick] (CA) edge node[left] {} (DA);
    \path [-, thick] (DA) edge node[left] {} (EA);
    \path [-, thick] (AB) edge node[left] {} (BB);
    \path [-, thick] (BB) edge node[left] {} (CB);
    \path [-, thick] (CB) edge node[left] {} (DB);
    \path [-, thick] (DB) edge node[left] {} (EB);
    \path [-, thick] (AC) edge node[left] {} (BC);
    \path [-, thick] (BC) edge node[left] {} (CC);
    \path [-, thick] (CC) edge node[left] {} (DC);
    \path [-, thick] (DC) edge node[left] {} (EC);
    \path [-, thick] (AD) edge node[left] {} (BD);
    \path [-, thick] (BD) edge node[left] {} (CD);
    \path [-, thick] (CD) edge node[left] {} (DD);
    \path [-, thick] (DD) edge node[left] {} (ED);
    \path [-, thick] (AE) edge node[left] {} (BE);
    \path [-, thick] (BE) edge node[left] {} (CE);
    \path [-, thick] (CE) edge node[left] {} (DE);
    \path [-, thick] (DE) edge node[left] {} (EE);
    \path [-, thick] (AA) edge node[left] {} (AB);
    \path [-, thick] (AB) edge node[left] {} (AC);
    \path [-, thick] (AC) edge node[left] {} (AD);
    \path [-, thick] (AD) edge node[left] {} (AE);
    \path [-, thick] (BA) edge node[left] {} (BB);
    \path [-, thick] (BB) edge node[left] {} (BC);
    \path [-, thick] (BC) edge node[left] {} (BD);
    \path [-, thick] (BD) edge node[left] {} (BE);
    \path [-, thick] (CA) edge node[left] {} (CB);
    \path [-, thick] (CB) edge node[left] {} (CC);
    \path [-, thick] (CC) edge node[left] {} (CD);
    \path [-, thick] (CD) edge node[left] {} (CE);
    \path [-, thick] (DA) edge node[left] {} (DB);
    \path [-, thick] (DB) edge node[left] {} (DC);
    \path [-, thick] (DC) edge node[left] {} (DD);
    \path [-, thick] (DD) edge node[left] {} (DE);
    \path [-, thick] (EA) edge node[left] {} (EB);
    \path [-, thick] (EB) edge node[left] {} (EC);
    \path [-, thick] (EC) edge node[left] {} (ED);
    \path [-, thick] (ED) edge node[left] {} (EE);
\end{tikzpicture}
        \caption{}
    \end{subfigure}
    \begin{subfigure}[]{0.5\columnwidth}
        \centering
\begin{tikzpicture}[every node/.style={transform shape}]
    \node[shape=circle,draw=black, fill=black] (AA) at (0,0) {};
    \node[shape=circle,draw=black, fill=black] (BA) at (1.5,0) {};
    \node[shape=circle,draw=black, fill=black] (CA) at (3,0) {};
    \node[shape=circle,draw=black, fill=black] (AB) at (0,1.5) {};
    \node[shape=circle,draw=blue, fill=blue] (BB) at (1.5,1.5) {};
    \node[shape=circle,draw=black, fill=black] (CB) at (3,1.5) {};
    \node[shape=circle,draw=black, fill=black] (AC) at (0,3) {};
    \node[shape=circle,draw=black, fill=black] (BC) at (1.5,3) {};
    \node[shape=circle,draw=black, fill=black] (CC) at (3,3) {};

    
    \path [->, draw=red, thick] (AA) edge node[left] {} (BA);
    \path [->, draw=red, thick] (BA) edge node[left] {} (CA);
    \path [->, draw=red, thick] (AB) edge node[left] {} (BB);
    \path [->, draw=red, thick] (BB) edge node[left] {} (CB);
    \path [->, draw=red, thick] (AC) edge node[left] {} (BC);
    \path [->, draw=red, thick] (BC) edge node[left] {} (CC);
    \path [->, thick] (AA) edge node[left] {} (AB);
    \path [->, thick] (AB) edge node[left] {} (AC);
    \path [->, thick] (BA) edge node[left] {} (BB);
    \path [->, thick] (BB) edge node[left] {} (BC);
    \path [->, thick] (CA) edge node[left] {} (CB);
    \path [->, thick] (CB) edge node[left] {} (CC);
\end{tikzpicture}
        \caption{}
    \end{subfigure}
    \caption{Template graphs for a Steerable CNN~(a) and a standard convolutional neural network~(b).
        In each case the output of the neuron is nonzero only for permutations that preserve the set of blue nodes
        and is invariant to permutation of the black nodes.
    }
    \label{fig:cnn_templates}
\end{figure}

\subsection{Steerable CNNs}
We first recover the steerable CNN architecture for the p4m group described in Reference~\cite{Cohen2017}. 
For concreteness, we will consider a steerable CNN constructed using a $3 \times 3$ filter; however,
the same construction can be applied to an arbitrary $k \times k$ filter.
To recover a $3 \times 3$ steerable CNN, we will use as our template a grid of size $5 \times 5$.  
Using this larger template allows us to easily express the aggregation of all the signals in the neuron's immediate receptive domain.
Note that this resembles the use of a star graph for MPNNs where we include both the ``input'' and the ``output'' nodes in the same template.

We first describe the output of the neuron
Each neuron outputs a function on $\Sbb_{25}$ that (a) is nonzero only for permutations that preserve the center $4 \times 4$ nodes and (b) is constant for all permutations of the black nodes amongst each other.  
Furthermore, the output is nonzero only for elements corresponding to the automorphism group of the $3 \times 3$ grid. This is the group $D_4$ which corresponds to rotations of the grid by 90 degrees and horizontal / vertical reflections.

These outputs then form the input of the neuron in the next layer.
Each neuron receives as input a function over $D_4$ associated with each $3 \times 3$ subgrid in Figure~\ref{fig:cnn_templates}a.
After narrowing and promotion, each of these function is embedded on group elements of $\Sbb_{25}$ that first send the grid to the indices $1,\ldots, 9$ and then applies the permutation corresponding to the appropriate element of $D_4$.
Finally, each neuron applies a convolution that combines subgrid group elements of similar orientations together using a convolution over the automorphism group of the template, which is also $D_4$.

This example highlights the importance of having a formalism capable of more complex methods of transferring information between neurons than merely copying over node or edge features.
If the input features had been sent to individual nodes and edges than we would have inadvertently averaged input signals over nodes or edges shared between incoming $3 \times 3$ graphs.

\subsection{Convolutional Neural Networks}
As discussed in the main text, convolutional neural networks have a notion of left-right and up-down.
This is information not present in a grid graph.
To hope to recover a CNN from any graph neural network architecture, we must therefore consider a richer graph embedding for the image.

For a one-dimensional CNN, we can recover a CNN by considering a directed graph where edges always point in one direction.
Two extend this to a two-dimensional CNN, we introduce two different types of edges: ``red'' edges that move horizontally from pixel to pixel in the image and ``black'' edges that move vertically.
We arbitrarily set the red edges to always point towards the right and the black edges to always point up.
Then, we construct a template using the same red and black edges (depicted in Figure~\ref{fig:cnn_templates}b).
Each template then picks out a single $k \times k$ subgrid in the image (again, we set $k=3$ for specificity in the discussion that follows).
To construct each neuron, we first extend our definition of automorphism to require that automorphism must also preserve color.
In this case the automorphism group of the template is the trivial group and we can take an arbitrary linear combination of all the pixels covered by our template.
This is the operation performed by a standard CNN filter.

This example gives insight into the role of the automorphism group in the Autobahn formalism.  
In the main text the automorphism group was defined as the set of all permutations that leaves the Adjacency matrix unchanged.
This choice was made simply because every graph has an adjacency matrix,
ensuring that the automorphism group would be well-defined for any graph.
However, if we know more about the structure of our graph, there is no reason that information cannot be included into the definition.
For instance, in this example we have also required that the permutations leave a collection of edge features unchanged as well (specifically edge color).
In other applications one might also further constrain the group by considering other graph attributes such as node features.

\section{Independence of representative for promotion.}

Uniqueness of $u$ and $s$ in the definition of promotion is a straightforward application of the following lemma.
\begin{lemma}
    Let $\tau$ and $t$ be elements of $\Sm$. Let $u, v \in \Sk$ and $s, q \in \Sbb_{m-k}$ be permutations such that
\begin{equation}
    \tau =  \acute{u} \grave{s} t = \acute{v} \grave{q} t
\end{equation}
Then $u = v$ and $s=q$.
\end{lemma}
\begin{proof}
    It follows directly from the assumption that 
    \begin{align*}
        &\acute{u} \grave{s} = \acute{v} \grave{q} \\
        \implies & \acute{u} \grave{s} \acute{v}^{-1} \grave{q}^{-1} = \mathcal{I}
    \end{align*}
    Since $\acute{v}^{-1}$ acts only on the first $k$ elements and $\grave{s}$ acts only on the last $k$, they commute and 
    \begin{equation*}
        \acute{u} \acute{v}^{-1} \grave{s} \grave{q}^{-1} = \mathcal{I}
    \end{equation*}
    Moreover, let $a= u v^{-1}$ and $b=s q^{-1}$.
    It follows from the definition of $\;\acute{}\;$ and $\;\grave{}\;$ that 
    \begin{equation}
        \acute{a} = \acute{u} \acute{v}^{-1} \text{ and } \grave{b} = \grave{s} \grave{q}^{-1}.
    \end{equation}
    implying
    \begin{equation}
        \acute{a} \grave{b} = \mathcal{I}.
    \end{equation}
    But this is only possible if $a$ is the identity element of $\Sk$ and $b$ is the identity element of $\Sbb_{m-k}$, which in turn implies $u=v$  nd $s=q$.
\end{proof}

\section{Proof of Equivariance for Autobahn}
Here, we prove that the Autobahn architecture obeys permutation equivariance.
For Autobahn, neural network inputs and activations are functions on subgroups acting on a subset of the graph's vertices
and global permutation of inputs can induce group operations in the associated subgroups.

Proving equivariance requires we describe how Autobahn networks transform when applying a global permutation to the input graph.
To do so, we recall from Subsection~4.2 that narrowing is the pseudoinverse of promotion.
Consequently, between every step in Autobahn
we can promote the activation of any neuron to $\Sn$ 
and then immediately narrow it back to the nodes in the neuron's local graph
without changing the network.
Doing this allows us to rewrite the steps for an Autobahn neuron \m{\neuron^\ell_j} 
as follows.
\begin{compactenum}
    \item[T1.1~] Narrow every activation \m{f^{\ell-1}_{s_z}} from $S_n$ to \m{\cbd{\sseq{a^z}{m_z}}}.
    \item[T1.2~] Further narrow the incoming activation \m{f^{\ell-1}_{s_z}} to the corresponding 
intersection to get \m{f^{\ell-1}_{s_z}\dn_{(\ssseq{b^z}{k_z})}}.   
    \item[T1.3~] Promote each \m{f^{\ell-1}_{s_z}\dn_{(\ssseq{b^z}{k_z})}} to $S_n$.
    \item[T2.1~] Narrow the results back to \m{(\ssseq{b^z}{k_z})}. 
    \item[T2.2~] Promote each of these to \m{(\sseq{a}{m})}\,: 
    \vspace{-5pt}
    \[ \tilde f_z=f^{\ell-1}_{s_z}\dn_{(\ssseq{b^z}{k_z})}\up^{(\ssseq{a}{m})}.
    \vspace{-8pt}
    \]
    \item[T2.3~] Promote \m{\tilde f_z} to $S_n$.
    \item[T3.1~] Narrow back down to \m{(\sseq{a}{m})}.
    \item[T3.2~] Apply a symmetric polynomial \m{S} to \m{\sseq{\tilde f}{p}}\,: 
    \vspace{-5pt}
    \[\h f=S(\sseq{\tilde f}{p}).
    \vspace{-8pt}
    \]
    \item[T3.3~] Promote $\h f$ to $S_n$.
    \item[T4.1~] Narrow back down to \m{(\sseq{a}{m})}.
    \item[T4.2~] Apply one or more (learnable) equivariant linear transformations \m{p_j}, 
        each of which is followed by a fixed pointwise nonlinearity \m{\xi}, to get the final output of the neuron. 
    \vspace{-5pt}
    \[f^\ell_j=\xi(p_j(\h f)).
\]
    \item[T4.3~] Promote every $f^\ell_j$ to $S_n$.
\end{compactenum}
Here, we have rewritten Autobahn so that steps T1, T2, T3, and T4 all map functions on $\Sn$ to functions on $\Sn$.
Consequently, we can show that each of these steps is equivariant to permutations in $\Sn$, making the network to equivariant to permutation as whole.

\subsection{Action of permutations on narrowing and promotion}
Before we prove equivariance, we first consider how a permutation of a local graphs affects the arbitrary orderings chosen when narrowing and promoting.
Consider narrowing a function on $m$ vertices onto a subset of $k$ vertices, $\cbr{\sseq{i}{k}}\subset \cbr{1, \ldots, m}$.
Narrowing first applies a permutation that sends the vertices indexed by $\cbr{\sseq{i}{m}}$ to the first $k$ indices, $(1, \ldots, m)$.
Vertex $i_1$ is sent to position $1$, vertex $i_2$ is sent to position $2$, and so forth.
However, this implicitly orders the vertices:
If we had initially listed the vertices in a different order then they would have been sent to different positions in the set $(1\ldots,m)$.
Consequently, when applying a permutation $\pi$ to a local graph graph with $m$ nodes, it is not enough to merely consider narrowing from $\cbr{\pi(i_1),\ldots,\pi(i_m)}$ to $(1,\ldots,m)$ as we have no guarantee of recovering the same arbitrary ordering when considering a permuted copy of the graph.
Rather, we must also ensure that our network is unaffected by subsequent permutation $p$ of the labels $(1, \ldots, m)$ caused by making a different arbitrary choice in order for the permuted graph.
Throughout this section, we will extend the $\; \acute{} \;$ and $\; \grave{} \;$ notation to both maps from $\Sm$ to $\Sn$ or $\Sk$ to $\Sn$ as necessary
and trust that the precise domain and ranges of the maps will be clear from context.

Let $f$ be an activation and $u\in\Sk$, $s \in \Sbb_{m-k}$, and $t \in \Sm$ be permutations as in Subsection~5.2 of the main text.
Upon applying a permutation $\pi$ to a neuron's local graph, the neuron narrows the permuted activation $T_\pi f$ to the function
\begin{equation}
    (T_\pi) f \dn_{(\pi(i_{p(1)}), \pi(i_{p(2)}), \ldots, \pi(i_{p(k)}))}(u)
        = {(n-k)!}^{-1}\hspace{-5pt} 
        \sum_{s \in \Sbb_{m-k}} T_\pi f(\acute{u} \grave{s} \tau).
\end{equation}
where $\tau$ is an arbitrary permutation that sends $\pi(i_1)$ to $p(1)$, $\pi(i_2)$ to $p(2)$, etc.
Moreover, since $t$ and $\tau \pi$ send the same vertices to the (unordered) sets $\cbr{1, \ldots, k}$ and $\cbr{k+1, \ldots, m}$ there must exist 
a permutation $a \in \Sbb_{k-m}$ such that
\begin{equation}
    \acute{p}  \grave{a} t =   \tau \pi.
\end{equation}
and we can consequently write 
\begin{align}
    T_\pi f \dn_{(\pi(i_{p(1)}), \pi(i_{p(2)}), \ldots, \pi(i_{p(k)}))}(u)
        &= {(n-k)!}^{-1}\hspace{-5pt} 
        \sum_{s \in \Sbb_{m-k}} T_\pi f(\acute{u} \grave{s} \tau). \nonumber \\
        &= {(n-k)!}^{-1}\hspace{-5pt} 
        \sum_{s \in \Sbb_{m-k}} T_\pi f(\acute{u} \grave{s} \acute{p} \grave{a} t \pi^{-1} ). \nonumber \\
        &= {(n-k)!}^{-1}\hspace{-5pt} 
        \sum_{s \in \Sbb_{m-k}} f(\acute{u} \grave{s} \acute{p} \grave{a} t \pi^{-1} \pi ). \nonumber \\
        &= 
        \sum_{s \in \Sbb_{m-k}} f(\acute{u} \acute{p} \grave{s} \grave{a} t). \nonumber \\
        &= 
        f \dn_{(\pi(i_{p(1)}), \pi(i_{p(2)}), \ldots, \pi(i_{p(k)}))}(u p)  \nonumber \\
        &= 
        T_p f \dn_{(\pi(i_{p(1)}), \pi(i_{p(2)}), \ldots, \pi(i_{p(k)}))}(u). \label{eq:perm_on_narrowing}
\end{align}
Similarly, promoted functions transform as
\begin{align}
    g\up^{\left(\pi\left(i_{p(1)}\right), \ldots, \pi\left(i_{p(k)}\right)\right)}(\alpha) &=     \begin{cases}
        g(u) &\text{ if } \exists u \in \Sk, s \in \Sbb_{m-k} \text{ s.t. } \alpha = \acute{u} \grave{s} \tau \\
        0 & \text{ otherwise.}
    \end{cases} \\
    &=     \begin{cases}
        g(u) &\text{ if } \exists u \in \Sk, s \in \Sbb_{m-k} \text{ s.t. } \alpha = \acute{u} \acute{p} \grave{s} \grave{a} t \pi^{-1} \\
        0 & \text{ otherwise.}
    \end{cases} \\
\implies
    T_p g\up^{\left(\pi\left(i_{p(1)}\right), \ldots, \pi\left(i_{p(k)}\right)\right)} &=  
    \left(T_\pi  g\right) \up^{\left( \sseq{i}{k}\right)} \label{eq:perm_on_promotion}
\end{align}

\subsection{Proof of equivariance for individual sublayers}
We now prove that each of the individual sublayers T1-T4 obey equivariance.

\subsubsection{T1 is equivariant}
Since we are applying narrowing twice, we must deal with two sets of ordered vertices:
$(i_1, \ldots, i_m)$ and the ordered subset to which we are narrowing, $(j_1, \ldots, j_k)$.

Our entire graph will be acted on by a permutation $\rho$.
We let $b\in\Sn$ be an arbitrary permutation that sends the ordered set $(i_1, \ldots, i_m)$ to the first $m$ positions and let $\beta\in\Sn$ be a permutation that sends $(\rho(i_1), \ldots, \rho(i_m))$ to $\left(\pi(1), \dots, \pi(m)\right)$ for some (unknown) $\pi$.
The permutations $b$ and $\beta$ play the same role as $t$ and $\tau$ when narrowing from $\cbr{1, \ldots, n}$ to $\left(i_1, \ldots, i_m\right)$.

To aid the reader, we have summarized how global permutations affect the various subsets involved in narrowing and promotion in~\ref{fig:commutative_diagram}.
\begin{figure}[t]
    \centering
    \begin{tikzcd}
        \left(\ldots, 5, 3, \dots, 2, 1, \ldots, 4 \ldots\right)
            \arrow[r, shift left, "b"] \arrow[d, shift left, 
            "\rho"
        ] 
        & \left(1, 2, 3, 4, 5 , \ldots   \right)  
            \arrow[d, shift left, 
            "\acute{\pi} \grave{\alpha}"
        ]  \arrow[r, shift left, "\acute{t}"] \arrow[l, shift left, "b^{-1}"]
        & \left(2,4, 3, \ldots\right) 
        \arrow[d, shift left, "\acute{p}\acute{\grave{a}}"]  \arrow[l, shift left, "\acute{t}^{-1}"]
 \\
        \left(\ldots, 4, 3, \dots, 5, \ldots 2, 4, 1 \right) 
             \arrow[r, shift left, "\pi"] \arrow[u, shift left, 
                 "\rho^{-1}"
            ]
        &  \left( 5, 4, 1, 3, 2, \ldots \right)\arrow[r, shift left, "\tau"] \arrow[l, shift left, "\pi^{-1}"] \arrow[u, shift left, 
            "\acute{\pi}^{-1} \grave{\alpha}^{-1}"
        ]
        &
         \left(4, 2,3, \ldots, \right) 
         \arrow[l, shift left, "\tau^{-1}"] \arrow[u, shift left, "\acute{p}^{-1}\acute{\grave{a}}^{-1}"]
    \end{tikzcd}
    \caption{Commutative diagram depicting the permutations involved in T1 and T2.
             For concreteness, we have set $k=3$ and $m=5$.
             Our local graph is defined on vertices vertices 1 through 5,
             and here we are narrowing to vertices $j_1=2$, $j_2=4$, and $j_3=2$.}
    \label{fig:commutative_diagram}
\end{figure}

We now seek to prove equivariance:
\begin{align}
    &\left(
        \left(
                (T_\rho f)
                    \dn_{\left(\rho\cbd{i_{\pi(1)}}, \ldots, \rho\cbd{i_{\pi(m)}}\right)}
        \right)
            \dn_{ 
                    \left( 
                        \pi \left( j_{p(1)} \right), 
                        \ldots, 
                        \pi \left( j_{p(k)} \right)  
                    \right)
                }
    \right)
    \up^{
        \left( 
            \rho\cbd{i_{j_{p(1)}}}
            \ldots, 
            \rho\cbd{i_{j_{p(k)}}}
        \right)
    } \nonumber \\
    &\qquad= 
    T_\rho
        \left(
            \left(
                \left(
                    f\dn_{\left(\sseq{i}{m}\right)}
                \right)
                    \dn_{\left(\sseq{j}{k}\right)}
            \right)
                \up^{\left(i_{j_1},\ldots,i_{j_k}\right)} 
        \right)
\end{align}
Repeatedly applying~\eqref{eq:perm_on_narrowing}, we have
\begin{align}
    &\left(
        \left(
                (T_\rho f)
                    \dn_{\left(\rho\cbd{i_{\pi(1)}}, \ldots, \rho\cbd{i_{\pi(m)}}\right)}
        \right)
            \dn_{ 
                    \left( 
                        \pi \left( j_{p(1)} \right), 
                        \ldots, 
                        \pi \left( j_{p(k)} \right)  
                    \right)
                }
    \right)
    \up^{
        \left( 
            \rho\cbd{i_{j_{p(1)}}}
            \ldots, 
            \rho\cbd{i_{j_{p(k)}}}
        \right)
    }
    \nonumber \\
    &\qquad= 
    \left(
        \left(
            T_\pi \left(f \dn_{\cbd{\sseq{i}{m}}}\right)
        \right)
            \dn_{ 
                    \left( 
                        \pi \left( j_{p(1)} \right), 
                        \ldots, 
                        \pi \left( j_{p(k)} \right)  
                    \right)
                }
    \right)
    \up^{
        \left( 
            \rho\cbd{i_{j_{p(1)}}}
            \ldots, 
            \rho\cbd{i_{j_{p(k)}}}
        \right)
    }
    \\
    &\qquad= 
    \left(
        T_p 
        \left(
            \left(f \dn_{\cbd{\sseq{i}{m}}}\right)
            \dn_{\left(\sseq{j}{k}\right)}
        \right)
    \right)
    \up^{
        \left( 
            \rho\cbd{i_{j_{p(1)}}}
            \ldots, 
            \rho\cbd{i_{j_{p(k)}}}
        \right)
    } \\
    &\qquad=
    T_\rho
    \left(
        \left(
            \left(f \dn_{\cbd{\sseq{i}{m}}}\right)
            \dn_{\left(\sseq{j}{k}\right)}
        \right)
    \right)
    \up^{
        \left( 
            i_{j_{1}}
            \ldots, 
            i_{j_{k}}
        \right)
    }
\end{align}
where the last line follows by the same argument as~\eqref{eq:perm_on_promotion}.

\subsubsection{T2 is equivariant}
We seek to prove that 
\begin{align}
    &\left(
        \left(
                (T_\rho f)
                \dn_{
                    \left( 
                        \rho\cbd{i_{j_{p^{-1}(1)}}}
                        \ldots, 
                        \rho\cbd{i_{j_{p^{-1}(k)}}}
                    \right)
                } 
        \right)
        \up^{
            \left(
                \pi \left( j_{p(1)} \right), 
                \ldots, 
                \pi \left( j_{p(k)} \right)  
            \right)
        }
    \right)
        \up^{\left(\rho\left(i_{(1)}\right), \ldots, \rho\left(i_{(m)}\right)\right)}
        \nonumber \\
    &\qquad= 
    T_\rho
        \left(
            \left(
                \left(
                    f
                        \dn_{\left(i_{j_1},\ldots,i_{j_k}\right)} 
                \right)
                    \up^{\left(\sseq{j}{k}\right)}
            \right)
            \up^{\left( i_1, \ldots, i_m \right)} 
        \right)
\end{align}
The proof proceeds similarly to the proof for T1.
We have
\begin{align}
    &\left(
        \left(
                (T_\rho f)
                \dn_{
                    \left( 
                        \rho\cbd{i_{j_{p(1)}}}
                        \ldots, 
                        \rho\cbd{i_{j_{p(k)}}}
                    \right)
                } 
        \right)
        \up^{
            \left(
                \pi \left( j_{p(1)} \right), 
                \ldots, 
                \pi \left( j_{p(k)} \right)  
            \right)
        }
    \right)
        \up^{\left(\rho\left(i_{\pi(1)}\right), \ldots, \rho\left(i_{\pi(m)}\right)\right)}
        \nonumber \\
    &\qquad= 
    \left(
        \left(
            T_p
            \left(
                f
                \dn_{\left(i_{j_1},\ldots,i_{j_k}\right)} 
            \right)
        \right)
        \up^{
            \left(
                \pi \left( j_{p(1)} \right), 
                \ldots, 
                \pi \left( j_{p(k)} \right)  
            \right)
        }
    \right)
        \up^{\left(\rho\left(i_{\pi(1)}\right), \ldots, \rho\left(i_{\pi(m)}\right)\right)}
        \\
    &\qquad= 
    \left(
        T_\pi
        \left(
            \left(
                f
                \dn_{\left(i_{j_1},\ldots,i_{j_k}\right)} 
            \right)
            \up^{
                \left( \sseq{j}{k} \right)
            }
        \right)
    \right)
        \up^{\left(\rho\left(i_{\pi(1)}\right), \ldots, \rho\left(i_{\pi(m)}\right)\right)}
        \\
    &\qquad= 
        T_\rho
            \left(
                \left(
                    \left(
                        f
                            \dn_{\left(i_{j_1},\ldots,i_{j_k}\right)} 
                    \right)
                        \up^{\left(\sseq{j}{k}\right)}
                \right)
                \up^{\left( i_1, \ldots, i_m \right)} 
            \right)
\end{align}

\subsubsection{T3 is equivariant}
To prove that applying a symmetric polynomial and applying convolution over a graph's automorphism group preserves equivariance, we require the following lemma.
\begin{lemma}\label{lem:internal_equivariance}
    Let $A$ be an $\Sm$ equivariant operator.
    Then,
    \begin{align}
    \left(
        A \left(
            (T_\rho f)
                \dn_{\left(\rho(i_{\pi(1)}), \ldots, \rho(i_{\pi(k)})\right)}
        \right)
    \right)
        \up^{\left(\rho(i_{\pi(1)}), \ldots, \rho(i_{\pi(k)})\right)} 
        &= T_\rho f
    \left(
        \left(
            A
            \left( 
                f \dn_{\left( i_1,\ldots, i_k \right)}
            \right)
        \right)
            \up^{\left(i_1, \ldots, i_k\right)}
    \right)
    \end{align}
\end{lemma}
\begin{proof}
    Applying \eqref{eq:perm_on_narrowing}, the definition of equivariance, and \eqref{eq:perm_on_promotion} we have
    \begin{align}
        &\left(
            A \left(
                (T_\rho f)
                    \dn_{\left(\rho(i_{\pi(1)}), \ldots, \rho(i_{\pi(k)})\right)}
            \right)
        \right)
            \up^{\left(\rho(i_{\pi(1)}), \ldots, \rho(i_{\pi(k)})\right)} \nonumber \\
    &\qquad=
        \left(
            A \left(
                T_\pi \left( f\dn_{\sseq{i}{m}} \right)
            \right)
        \right)
            \up^{\left(\rho(i_{\pi(1)}), \ldots, \rho(i_{\pi(k)})\right)} \nonumber \\
    &\qquad=
        \left(
            T_\pi \left(
                A \left( f\dn_{\sseq{i}{m}} \right)
            \right)
        \right)
            \up^{\left(\rho(i_{\pi(1)}), \ldots, \rho(i_{\pi(k)})\right)} \nonumber \\
    &\qquad=
        T_\rho f
            \left(
                \left(
                    A
                    \left( 
                        f \dn_{\left( i_1,\ldots, i_k \right)}
                    \right)
                \right)
                    \up^{\left(i_1, \ldots, i_k\right)}
            \right)
    \end{align}
\end{proof}

To prove equivariance of T3, it is therefore enough to prove that application of the symmetric polynomial is $\Sm$-equivariant.
However, the output of the polynomial is invariant by definition,
and it is well-known that the products are group-equivariant~\cite{SphericalCNN2018arxiv}.
Consequently, the symmetric polynomials obey equivariance.

\subsubsection{T4 is equivariant}
Equivariance of T4 follows directly from 
Lemma~\ref{lem:internal_equivariance}  and the fact the neuron described by Algorithm~1 is equivariant,
as shown in Section~\ref{sec:neuron_equivariance}.

\section{Expressivity of Autobahn}
Here, we give results directly comparing the expressivity of specific Autobahn architectures to other graph neural networks.

\subsection{Recovery of k'th-order Networks}\label{ssec:kth_order_comparison}

Here, we show that
\begin{enumerate}
    \item The expressivity of an Autobahn network whose largest template has $k$ nodes is bounded above by the expressivity of a $k$'th order network as described in~\citep{maron2018invariant}.
    \item There exists an Autobahn network that achieves this bound.
\end{enumerate}

\subsubsection{Group-theoretic description of k'th order networks}\label{sssec:kth_order}
To facilitate our discussion, we first describe the $k$'th order networks from a group theoretic point of view.
Here, the $k$'th order network can be described as functions on the quotient spaces $\Sn / \Sbb_{n-k}$, $\Sn / \Sbb_{n-k+1}$, etc.
The different orbits of the permutation group on the $k$'th order tensor correspond to different quotient spaces: for instance, the central diagonal 
Note that in general, a function on $\Sn / \Sbb_{n-m}$ can be identified with a function on $\Sn$ that obeys the following property
\begin{equation}
    g(u) =  g ( u \grave{s})\; \forall \; s \in \Sbb_{n-m}.
    \label{eq:Sn_level_sets}
\end{equation}
The value of such a function depends only on which nodes $u$ sends to the first $m$ positions:
the indices of these nodes correspond, in order, to the indices in a tensor representation of the data.
Note that the original formulation of the $k$'th order network is given in terms of tensors with indices corresponding to individual nodes;
the two formulations can be interconverted as described in Section~\ref{sec:group_activations}.
The action of the network itself can be fully described using the general theory of group equivariant networks described in~\citep{KondorTrivedi2018}.
However, here it will be easier to describe the network as the composition of three operations.
\begin{enumerate}
    \item Convolution over all possible permutations of the first $m$ indices
        \begin{equation}
            g \star h (u) = \sum_{s\in \Sk} g(\acute{s} u) h(s^{-1})
            \label{eq:kth_order_conv}
        \end{equation}
        Here $g$ is a function obeying~\eqref{eq:Sn_level_sets} and $h$ is a function from $\Sk \to \Ccal$.
        It is easy to see that the output of the convolution
        is on the same quotient space as $g$.
        Note that by setting $h$ to a function that is $1$ for a single group element and zero otherwise and convolving, it is possible to use~\eqref{eq:kth_order_conv} to translate $g$ by a permutation that permutes the first $k$.
    \item Moving a function onto a smaller homogeneous space by averaging over all possible permutations of the other indices.
        If we look at the corresponding functions on $\Sn$, this corresponds to performing the following sum.
        \begin{equation}
            g'(u) = \sum_{s\in \Sbb_{n-k}}  g(\acute{s} u)
             \label{eq:reduction_in_kth_order}
        \end{equation}
        This corresponds to averaging over the last $n-k$ indices.
   \item 
       Moving a function from a smaller homogeneous space $\Sn / \Sbb_{n-m}$ to a larger homogeneous space $\Sn / \Sbb_{n-k}$.
       The precise characterization of this function depends on how we view the inputs and outputs.
       Writing them as $m$'th order and $k$'th order tensors $\gamma$ and $\gamma'$ respectively, this corresponds to setting 
       \begin{equation}
           \gamma'_{i_1, i_2, \ldots, i_m, \ldots, i_k} = \gamma_{i_1, i_2, \ldots, i_m}.
       \end{equation}
       However, viewed as a function on $\Sn$, this leaves the function unchanged.
\end{enumerate}

\subsubsection{Embedding Autobahn Activations in a k'th order network}
It is sufficient to prove that the $k$'th order network can reproduce Autobahn for templates of size $k$ and a $k$'th order signal, as we can trivially
treat a lower-order signal as a $k$'th order signal that has the same value for many choices of index.
First, we show that the $k$'th order network is capable of representing the Autobahn activation.
Consider an Autobahn activation $f_j$ whose local graph is denoted $\rfd_j$.  
Using promotion, we can embed the activation as a function on $\Sn$ as follows.
Let $t$ be an arbitrary permutation that moves the indices of the $j$'th local graph to $1, \ldots, k$.
\begin{equation}
    f_j\up^{\left(1,\ldots,n\right)} (\alpha) = 
    \begin{cases}
        f_j(u) &\text{ if } \exists u \in \Sk, s \in \Sbb_{m-k} \text{ s.t. } \alpha = \acute{u} \grave{s} t \\
        0 &\text{otherwise}. 
    \end{cases}
\end{equation}
More specifically, we note that since the general theory of group equivariant networks says the activation of an Autobahn neuron (here arbitrarily chosen to be neuron $j$) can be written as a function over $\Aut(\rfd_j)$, and will consequently have an even more structured sparsity pattern.
Specifically, we define the $\; \hat{} \;$ symbol, similar to the $\; \grave{} \;$, that maps elements of $\Aut(\rfd_j)$ to $\Sn$ such that for $u \in \Aut(\rfd_j)$,
\begin{align}
    \hat{u}(i) &= u(i) \quad  \forall \; 1 \leq i \leq k \\ 
    \hat{u}(i) &= i \quad \forall \; i > k.
\end{align}
Note that if a permutation can be written as $\hat{u}$, it can also be written using the  $\: \acute{} \;$ symbol using Cayley's theorem.
Consequently, for any given choice of $t_j$ that sends the indices here exists a permutation in $v_j\in \Sk$ such that 
\begin{equation}
    f_j\up^{\left(1,\ldots,n\right)} (a) = 
    \begin{cases}
        f_j(u) &\text{ if } \exists u \in \Aut(\rfd_j), s \in \Sbb_{m-k} \text{ s.t. } a = \hat{u} \acute{v}_j \grave{s} t_j \\
        0 &\text{otherwise}. 
        \label{eq:Automorphic_embedding}
    \end{cases}
\end{equation}
Note this embedding is (trivially) injective.

We now extend this embedding to the multiple neuron case.
Without loss of generality, we assume that each subgraph isomorphic to a template corresponds to single neuron. (If two neurons are operating on the same subgraph they can be treated as a single neuron operating on more channels.)
We can always sort the neurons based on their isomorphism class without breaking permutation equivariance. 
Consequently we need only consider neurons that share a template when constructing an embedding: neurons whose local graphs are not isomorphic can be embedded separately, and then concatenated along the channel index.
Denoting the set of neurons that share the template by $\neuronT$,
we can embed their activations in $\Sn / \Sbb_{n-k}$ as
\begin{equation}
    f^{emb}(a) = \sum_{j=1}^{\neuronT} f_j\up^{\left(1,\ldots,n\right)}(a)
    \label{eq:embedding}
\end{equation}
Since each subgraph isomorphic to a template corresponds to a single neuron, for every value of (a) there is at most one promoted activation that is nonzero.
Consequently, this embedding is also injective.

Finally, we note that Autobahn explicitly uses information about the location of subgraphs in the networks.
This can be transferred into a $k$'th order network by embedding functions that are $1$ on a neuron, or on the intersection between neurons, in the same manner as in~\eqref{eq:embedding}.

\subsubsection{Performing Automorphic Convolution in a k'th order network}
We now show that the $k$'th order network is capable of performing operations on the embedded signal that correspond to the core operations of Autobahn: convolution over the Automorphism group, narrowing, and promotion.
Let $f_j$ and $g$ be a functions on $\Sk$ that are nonzero only for elements of the Automorphism group of $\rfd_j$.
We first observe that 
\begin{equation}
    \left(f_j \ast g\right)\up^{\left(1,\ldots,n\right)} (a) =  f_j \up^{\left(1,\ldots,n\right)} \star g (a)
\end{equation}
To show this, we note that if the top condition in~\eqref{eq:Automorphic_embedding} holds, we have that 
\begin{align}
    \left(f_j \ast g\right)\up^{\left(1,\ldots,n\right)} (a) 
        =& \sum_{w\in\Aut(\rfd_j)} f_j(w^{-1} u) g(w) \\
        =& \sum_{w\in\Aut(\rfd_j)} f_j \up^{\left(1,\ldots,n\right)} (\hat{w}^{-1} \hat{u} \acute{v}_j \grave{s} t_j) g (w)  \\
        =& \sum_{w\in\Sk} f_j \up^{\left(1,\ldots,n\right)} (\acute{w}^{-1} \hat{u} \acute{v}_j \grave{s} t_j) g (w)  \\
        =& f_j \up^{\left(1,\ldots,n\right)} \star g (a)
\end{align}
Similarly, if the bottom condition holds then 
\begin{align}
    \left(f_j \ast g\right)\up^{\left(1,\ldots,n\right)} (a)  =& 0 \\
    =& \sum_{s\in \Sk} f_j \up^{\left(1,\ldots,n\right)} (\acute{s} a) g(s^{-1}) \\
    =& f_j \up^{\left(1,\ldots,n\right)} \star g (a).
\end{align}
The second line follows from the fact that if $f_j (\acute{s} a)$ being nonzero implies that the first condition in ~\eqref{eq:Automorphic_embedding} holds instead of the second, contradicting our assumption.

We now return to our embedding of the Autobahn activations.  Embedding the output of each neurons convolution over their respective local graph's automorphism group,
\begin{align}
    \sum_{j=1}^{\neuronT} \left(f \ast g\right)_j\up^{\left(1,\ldots,n\right)} (a) 
    =& \sum_{j=1}^{\neuronT}  \sum_{s\in \Sk} f_j \up^{\left(1,\ldots,n\right)}(\acute{s} a)  g (s^{-1}) \\
    =& \left(\sum_{j=1}^{\neuronT}  \sum_{s\in \Sk} f_j \up^{\left(1,\ldots,n\right)}(\acute{s} a)\right)  g (s^{-1}) \\
    =& \left(\sum_{j=1}^{\neuronT}  f_j \up^{\left(1,\ldots,n\right)} \right) \star g (a)
\end{align}
Consequently, convolutions over the automorphism groups of individual neurons can be written using convolutions of the form of~\eqref{eq:kth_order_conv} on the embedded signal.
\subsubsection{Performing Narrowing in a k'th order network}
We first establish two useful identities.  We first note that 
\begin{align}
    g\dn_{(i_1, \ldots, i_k)} \dn_{(j_1, \ldots, j_m)} = g\dn_{(i_{j_1}, \ldots i_{j_m})}
    \label{eq:double_narrow}
\end{align}
and
\begin{equation}
    g\dn_{(i_1, \ldots i_k)} \up^{(1,\ldots,n)}(a) 
    =
        \begin{cases}
            &\sum_{w \in S_{n-m}} g (\acute{u} \grave{w} t)\text{ if } \exists u \in \Sk, s \in \Sbb_{n-k} \text{ s.t. } a = \acute{u} \grave{s} t  \\
            &0\text{ otherwise.}
        \end{cases}
\end{equation}
This latter identity can be written more compactly by introducing an indicator function $\mathds{1}$ that is 1 if the condition holds and 0 otherwise.
Since 
\begin{equation}
    \exists u \in \Sk, s \in \Sbb_{n-k} \text{ s.t. } a = \acute{u} \grave{s} t \iff \left\{a(i_1), \ldots, a (i_k)\right\} = \left\{ 1, \ldots, k   \right\}
\end{equation}
we can write 
\begin{equation}
    g\dn_{(i_1, \ldots i_k)} \up^{(1,\ldots,n)}(a) = \left(\sum_{w \in S_{n-m}} g( \grave{w} a ) \right) \mathds{1}_{\Ai} (a)
\end{equation}
where
\begin{equation}
    \Ai =\left\{\alpha \in \Sn | \alpha(i_1), \ldots, \alpha(i_k)\right\} = \left\{ 1, \ldots, k   \right\}.
    \label{eq:Ai_permutations}
\end{equation}
This set is preserved by permutation of the other $N-k$ indices, and consequently
\begin{equation}
    g\dn_{(i_1, \ldots i_k)} \up^{(1,\ldots,n)} (a) = \sum_{w \in S_{n-m}} g( \grave{w} a ) \mathds{1}_{\Ai} (\grave{w} a)
    \label{eq:narrow_then_promote}
\end{equation}
Combining~\eqref{eq:double_narrow} and~\eqref{eq:narrow_then_promote}, we have
\begin{align}
    \left(\sum_{i=1}^\neuronT  f_i \dn_{({j_1},\ldots,{j_m})}\right)\up^{(1,\ldots,n)}(a) 
    =& \sum_{i=1}^\neuronT  f_i \dn_{({j_1},\ldots,{j_m})} \up^{(1,\ldots,n)}(a)  \\
    =&  \sum_{i=1}^\neuronT f_i \up^{(1,\ldots,n)} \dn_{(i_1,\ldots,i_k)} \dn_{(j_1,\ldots,j_m)} \up^{(1,\ldots,n)}(a) \\
    =&  \sum_{i=1}^\neuronT f_i \up^{(1,\ldots,n)} \dn_{(i_{j_1},\ldots,i_{j_m})}  \up^{(1,\ldots,n)} (a) \\
    =&  \sum_{i=1}^\neuronT \sum_{s\in\Sbb_{n-m}}f_i \up^{(1,\ldots,n)} (\grave{s} a) \mathds{1}_{\Aij} (\grave{s} a)
\end{align}
where is the set of permutations such that
\begin{equation}
    \Aij = \left\{\alpha \in \Sn | \alpha(i_{j_1}), \ldots, \alpha(i_{j_m})\right\} = \left\{ 1, \ldots, m \right\}.
    \label{eq:Aij_permutations}
\end{equation}
Here the $i$ indices are chosen from an arbitrary neuron containing the graph intersections;
there is no dependence on $i$ on the left-hand side because the actual values of $i_{j_1}$ are independent of the neuron used.
Summing both sides over the $\meuronT$ possible intersections to which we are narrowing our activations gives
\begin{align}
    \sum_{j=1}^\meuronT \left(\sum_{i=1}^\neuronT  f_i \dn_{({j_1},\ldots,{j_m})}\right)\up^{(1,\ldots,n)}(a) 
    =&  \sum_{j=1}^\meuronT \sum_{i=1}^\neuronT \sum_{s\in\Sbb_{n-m}}f_i \up^{(1,\ldots,n)} (\grave{s} a) \mathds{1}_{\Aij} (\grave{s} a) \\
    =&    \sum_{s\in\Sbb_{n-m}}\left( \sum_{i=1}^\neuronT f_i \up^{(1,\ldots,n)} (\grave{s} a) \right)   \left(\sum_{j=1}^\meuronT \mathds{1}_{\Aij} (\grave{s} a) \right)
\end{align}
Consequently, we can write narrowing as an operation on a $k$'th order network by first applying an elementwise multiplication against a signal,
followed by an average over the remaining indices.

\subsubsection{Performing Promotion in a k'th order network}
To write promotion as an operation in a $k$'th order network, let $\left\{f_j\right\}$ a function being promoted from 
a subgraph on indices $i_{j_1}, \ldots, i_{j_m}$ to $i_1, \ldots, i_k$.
We observe that
\begin{equation}
    g \dn_{(i_{j_1},\ldots,i_{j_m})} \up^{(i_1, \ldots, i_k)} \up^{(1,\ldots,n)}(a) = 
    \sum_{w \in S_{n-m}} g(\grave{w} a) \mathds{1}_{\Ai}(a) \mathds{1}_{\Aij}(a).
\end{equation}
Now, we consider all $\meuronT$ functions being promoted from a set of isomorphic subgraphs.
We have
\begin{align}
    \sum_{j=1}^{\meuronT} f_j \up^{(i_1, \ldots, i_k)} \up^{(1,\ldots,n)} (a)
    =& \sum_{j=1}^{\meuronT} f_j \up^{(1, \ldots, n)} (a) \dn^{(i_{j_1}, \ldots, i_{j_m})} \up^{(i_1, \ldots, i_k)} \up^{(1,\ldots,n)} \\
    =& \sum_{j=1}^{\meuronT} f_j \up^{(1, \ldots, n)} (a) \mathds{1}_{\Ai}(a) \mathds{1}_{\Aij}(a) 
\end{align}
Summing over the target reference domain gives
\begin{align}
    \sum_{i=1}^{\neuronT} \sum_{j=1}^{\meuronT} f_j \up^{(i_1, \ldots, i_k)} \up^{(1,\ldots,n)} (a)
    = \sum_{i=1}^{\neuronT} \sum_{j=1}^{\meuronT} f_j \up^{(1, \ldots, n)} (a) \mathds{1}_{\Ai}(a) \mathds{1}_{\Aij}(a) 
\end{align}
Finally, we note that since each element of $f_j \up^{(1, \ldots, n)} (a)$ is nonzero only for a single $\mathds{1}_{\Aij}$, we can write
\begin{equation}
    \sum_{i=1}^{\neuronT} \sum_{j=1}^{\meuronT} f_j \up^{(i_1, \ldots, i_k)} \up^{(1,\ldots,n)} (a)
    = \left(\sum_{i=1}^{\neuronT} \sum_{j=1}^{\meuronT} f_j \up^{(1, \ldots, n)} (a)  \right) \left( \sum_{i=1}^{\neuronT} \sum_{j=1}^{\meuronT} \mathds{1}_{\Ai}(a) \mathds{1}_{\Aij}(a) \right)
    \label{eq:promotion_on_Sn}
\end{equation}
Consequently, promotion corresponds to multiplying the embedded activation with indicator functions on the sets of permutations that preserve the reference domains.

\subsubsection{Writing a k'th order network as an Autobahn}

To prove that this bound is tight, we construct an Autobahn network that can emulate the operations in Subsubsection~\ref{sssec:kth_order}.
To do this, we take as our templates the set of all edgeless graphs of size less than or equal to $k$.
Consequently, each neuron corresponds to one (ordered) set of $k$ or less nodes.
We will then associate each set of neurons to a homogeneous space of the corresponding size.  
As the automorphism group of these is simply the permutation group applied to the corresponding nodes, the network is able to trivially 
perform the convolution in ~\eqref{eq:kth_order_conv}.
To recover~\eqref{eq:reduction_in_kth_order} we apply narrowing to move the activation from all larger neurons of size $k$ to a smaller neuron of size $m$, $\neuron_l$.
We simplify this by only considering larger neurons whose reference domain strictly includes $\rfd_l$: while this is not necessary, it simplifies our construction.
In this case, it is easy to see that the narrowed activation corresponds to the sum over all possible strings of the remaining $k-m$ indices, which corresponds to~\eqref{eq:reduction_in_kth_order}.

Finally, promotion can be used to move an activation from a smaller homogeneous space to a larger one.  
This follows from~\eqref{eq:promotion_on_Sn}.  Since there is one neuron corresponding to each set of $m$-node graphs,
there is exactly one $f_l$ that is nonzero for any given of $a$.
Consequently, the value of the function for any given value is precisely the value of $f_l \up^{(1, \ldots, n)}$.

\subsection{Comparison with GSN}\label{ssec:gsn_comparison}

Both the GSN architecture and Autobahn directly featurize graphs using the isomorphism classes of 
subgraphs.  
However, whereas GSN uses the isomorphism classes to featurize an MPNN architecture, 
in Autobahn the neurons convolve directly over the corresponding automorphism group.
Consequently, it is natural to ask whether the automorphism group adds any additional 
flexibility compared to merely using isomorphism to construct features.
The following result answers in the affirmative.

\begin{theorem}
Consider a GSN network with a given set of subgraphs chosen to construct initial features. Next, consider an analogous Autobahn network where each layer’s neurons consist of (1) neurons that operate exactly the same way as the MPNN neurons in the GSN (2) Autobahn neurons that on the subgraphs used to featurize the GSN. The Autobahn network is at least as expressive as the GSN network. Moreover, there exists collections of GSN subgraphs such that the Autobahn network is strictly more expressive. 
\end{theorem}
    To prove that the Autobahn network is as expressive as the GSN, we show that it can imitate the behavior of the GSN. Since it has the same MPNN neurons as the GSN, it is enough to show the neurons operating on the subgraphs are able to learn a unique hash for each subgraph.  However, this is trivially true since each subgraph has its own weights.
    To show that the there exist subgraphs such that the Autobahn network is strictly more expressive, it is enough to give a single example.
    Consequently, we again consider templates corresponding to the edgeless graph of size $k$ or less. Since each node in an $N$-node graph is in exactly $N$ choose $k$ subgraphs isomorphic to the edgeless graph.  Each node receives exactly the same feature from the featurization graphs, so this does not improve GSN expressiveness over 1-WL. However, by the results above, the resulting Autobahn can recover a a $k$'th order network by simply ignoring the MPNN neurons.  Using results from~\citep{maron2018invariant} the network is consequently as powerful as the $k$-WL test, and is consequently strictly more powerful than the corresponding GSN.
Note that we cannot say that the Autobahn network is always strictly more expressive than the GSN network because of pathological counterexamples, e.g. setting all of the GSN’s subgraphs to be the subgraph of a single node.
In this case, the two networks trivially have the same expressivity.

\subsubsection{Connection with the Reconstruction Conjecture}

Previous work has analyzed the flexibility of neural networks in the light of the reconstruction conjecture\citep{kelly1957congruence,ulam1960collection}.
In the reconstruction conjecture, one takes a graph $\Gcal$  and forms the multiset of all subgraphs created by deleting a single node from $\Gcal$.
The conjecture then states that if two graphs have the same multiset then they must be isomorphic.
In~\citep{bouritsas2020improving}, it was noted that if the reconstruction conjecture held,  a GSN network featurized
on all subgraphs of size $N-1$ to be able to distinguish to isomorphic graphs.
However, the performance of the network if the reconstruction conjecture does not hold is unclear.

In contrast, Autobahn is able to reconstruct a graph from neurons operating on each of the $N-1$-node subgraphs even if the reconstruction conjecture does not hold.
We demonstrate this by outlining an explicit algorithm that achieves the reconstruction.
\begin{enumerate}
    \item The output of the first layer is a single edge feature that is 1 for all edges in the subgraph.
    \item Apply a single round of narrowing and promotion.  Since narrowing averages over all nodes that are outside of the intersection of the two sets,
        any edges that are outside the destination local graph are averaged to node features.
    \item Pick an arbitrary neuron.  Since only edges to nodes outside of the neuron's local graph were truncated, and there is exactly one node missing from the subgraph,
        connecting the missing node to any nodes that have nonzero node features reconstructs the graph.
\end{enumerate}
Note that this construction does not require any convolutions over the Automorphism group.  Instead, the Autobahn network is able to perform the reconstruction due to its ability to perform higher-order message passing.

\section{Architecture, hyper-parameter and computational details}

We provide some further details into the architecture, choice of hyper-parameters and training regime of our network.

\subsection{Architecture details}

We model our block specific (i.e. cycle / path) convolutions after standard residual convolutional networks \cite{he2016deep} with ReLU activations.
We refer the reader to our pytorch implementation for details of the implementation.

We perform a simple ablation study of the main components of the model on the Zinc-subset problem.
In particular, we study the impact of omitting: i) the cycle-based convolutions and ii) the paths shorter than the maximum length considered.
Validation loss\footnote{Note that for this particular dataset, validation losses tend to be higher than test losses, which is also observed in some other architectures such as HIMP.} results are reported in table~\ref{table:supplement-ablation}.

\sisetup{separate-uncertainty=true}
\begin{table}
    \centering
    \begin{tabular}{lr}
    \toprule
    Modification & Validation loss \\
    \midrule
    Original & \num{0.124 \pm 0.001} \\
    No cycles & \num{0.175 \pm 0.004} \\
    Only maximum length paths & \num{0.167 \pm 0.001} \\
    \bottomrule
    \end{tabular}
    \caption{Validation loss for modified architectures on the Zinc (subset) dataset.\label{table:supplement-ablation}}
\end{table}

\subsection{Hyper-parameter details}
Our hyper-parameters were chosen based on a combination of chemical intuition and cursory data-based selection, with some consideration towards computational cost. As our architecture specification is quite general, the number of hyper-parameters is potentially large.
In practice, we have restricted ourselves to tuning three parameters: a global network width parameter (which controls the number of channels in all convolutional and linear layers), a dropout parameter (which controls whether dropout is used and the amount of dropout), and the training schedule. The values used are specified in table~\ref{table:supplement-hyperparameters}.
The training schedule is set with a base learning rate of $0.0003$ at batch size 128 (and scaled linearly with batch size).
The learning rate is increased linearly from zero to the base rate during the specified number of warmup epochs, and is then piecewise-constant, with the value decaying by a factor of 10 after each milestone.
\begin{table}
    \centering
    \begin{tabular}{lrrrrr}
    \toprule
    Dataset & Channels & Dropout & Epochs & Warmup & Decay milestones \\
    \midrule
    Zinc (subset) & 128 & 0.0 & 600 & 15 & 150, 300 \\
    Zinc & 128 & 0.0 & 150 & 5 & 40, 80 \\
    MolPCBA & 128 & 0.0 & 50 & 5 & 35 \\
    MolHIV & 128 & 0.5 & 60 & 15 & N/A \\
    MolMUV &  64 & 0.0 & 30 & 5 & N/A \\
    \bottomrule
    \end{tabular}
    
    \caption{Hyper-parameter and training schedules used on each dataset.\label{table:supplement-hyperparameters}}
\end{table}

The lengths of the paths and cycles considered in the model are also hyper-parameters of the model.
We used the same values (cycles of lengths 5 and 6, and paths of lengths 3 to 6 inclusive) in all of our models.
We note that in molecular graphs, cycles of lengths other than 5 and 6 are exceedingly rare (e.g. in the Zinc dataset, cycles of lengths different from 5 or 6 appear in about 1\% of the molecules).
We evaluate different possibilities for the maximum length of paths to be considered in table~\ref{table:supplement-varying-path-length}, we observe that in general, both computational time and prediction performance increase with larger path lengths.

\begin{table}
    \centering
    
    \begin{tabular}{lrr}
    \toprule
    Path length & Validation Loss & Training Time \\
    \midrule
    4 & \num{0.140 \pm 0.000} & \SI{3}{\hour}\SI{15}{\minute} \\
    5 & \num{0.135 \pm 0.000} & \SI{4}{\hour}\SI{50}{\minute} \\
    6 & \num{0.124 \pm 0.000} & \SI{6}{\hour}\SI{50}{\minute} \\
    7 & \num{0.121 \pm 0.001} & \SI{9}{\hour}\SI{30}{\minute} \\
    \toprule
    \end{tabular}
    
    \caption{Performance and training time of model on Zinc (subset) as a function of maximum path length considered.
    \label{table:supplement-varying-path-length}}
\end{table}

\subsection{Computational details}

In a message passing graph neural network, computational time is typically proportional to the number of edges present in the graph.
On the other hand, our Autobahn network scales with the number of paths and cycles present in the graph.
An immediate concern may be that the number of such structures could be combinatorially large in the size of the graph.
In table~\ref{table:supplement-structure-count}, we show that, due to their tree-like structure, molecular graphs do not display such combinatorial explosion of number of sub-structures in practice.

The computational cost of our model scales roughly linearly with the total number of substructures under consideration.
In practice, for molecular graphs, selecting only paths of short lengths and cycles, we expect the computational cost to be on the same order of magnitude as standard graph neural networks.
We report the total amount of time (in GPU-hours) required for training each of our model in table~\ref{table:supplement-computational-cost}.
The training was performed on Nvidia V100 GPUs, and mixed-precision computation was used for all models except MolHIV were some gradient stability issue were encountered.
The two largest datasets (Zinc an MolPCBA) were trained on four GPUs, whereas the remaining datasets were trained on a single GPU.

\begin{table}[h]
    \centering
    \begin{tabular}{lrrrrrrrrrrrr}
    \toprule
     & & & \multicolumn{6}{c}{Paths} & \multicolumn{2}{c}{Cycles} \\
     \cmidrule(lr){4-9} \cmidrule(lr){10-11}
     Dataset & Nodes & Edges & 3 & 4 & 5 & 6 & 7 & 8 & 5 & 6 \\
     \midrule
     Zinc &\num{23.1}&\num{49.8}&\num{34.6}&\num{43.9}&\num{55.0}&\num{64.4}&\num{65.8}&\num{70.2}&\num{0.56}&\num{1.70} \\
     MolPCBA &\num{26.0}&\num{56.3}&\num{39.3}&\num{51.0}&\num{65.2}&\num{79.5}&\num{84.2}&\num{93.1}&\num{0.50}&\num{2.23} \\
     MolHIV &\num{25.5}&\num{54.9}&\num{39.2}&\num{52.1}&\num{68.9}&\num{87.2}&\num{97.2}&\num{111.5}&\num{0.34}&\num{2.01} \\
     MolMUV &\num{24.2}&\num{52.5}&\num{36.5}&\num{47.6}&\num{61.1}&\num{73.4}&\num{77.2}&\num{84.6}&\num{0.63}&\num{2.02} \\
     \bottomrule
    \end{tabular}
    \caption{
        Average count of structures in various datasets.
        \label{table:supplement-structure-count}
    }
\end{table}

\begin{table}[h]
    \centering
    \begin{tabular}{lrrr}
    \toprule
    Dataset & Samples & Samples / s / GPU & Training time (GPU-hours) \\
    \midrule
    Zinc (subset) & \num{6.0}\si{\mega} & \num{280} & \SI{6.8}{\hour} \\
    Zinc & \num{33.0}\si{\mega} & \num{215} & \SI{42.7}{\hour} \\
    MolPCBA & \num{21.9}\si{\mega} & \num{171} & \SI{35.5}{\hour} \\
    MolHIV & \num{2.5}\si{\mega} & \num{114} & \SI{6.1}{\hour} \\
    MolMUV & \num{2.8}\si{\mega} & \num{222} & \SI{3.5}{\hour} \\
    \bottomrule
    \end{tabular}
    
    \caption{
        Computational cost of training provided models.
        Samples denotes total number of gradients computed (i.e. number of epochs times number of observations in dataset).
        \label{table:supplement-computational-cost}
    }
\end{table}


\bibliography{gnn}
\bibliographystyle{unsrt}